\documentclass[lettersize,journal]{IEEEtran}
\usepackage{amsmath,amsfonts}
\usepackage{algorithm}
\usepackage{algorithmic}
\usepackage{array}
\usepackage[caption=false,font=normalsize,labelfont=sf,textfont=sf]{subfig}
\usepackage{textcomp}
\usepackage{stfloats}
\usepackage{subfig}
\usepackage{url}
\usepackage{verbatim}
\usepackage{graphicx}
\usepackage{float}
\usepackage{booktabs}
\usepackage{multirow}
\def\BibTeX{{\rm B\kern-.05em{\sc i\kern-.025em b}\kern-.08em
    T\kern-.1667em\lower.7ex\hbox{E}\kern-.125emX}}
\usepackage{balance}
\usepackage[colorlinks=true, linkcolor=blue, citecolor=blue, linktoc=all]{hyperref}
\IEEEtriggeratref{33}
\begin{document}
\title{A Novel Unified Extended Matrix for Graph Signal Processing: Theory and Application}
\author{Yunyan~Zheng, Zhichao~Zhang,~\IEEEmembership{Member,~IEEE}, and Wei~Yao
\thanks{This work was supported in part by the Open Foundation of Hubei Key Laboratory of Applied Mathematics (Hubei University) under Grant HBAM202404; and in part by the Foundation of Key Laboratory of System Control and Information Processing, Ministry of Education under Grant Scip20240121. \emph{(Corresponding author: Zhichao~Zhang.)}}
\thanks{Yunyan~Zheng and Wei~Yao are with the School of Mathematics and Statistics, Nanjing University of Information Science and Technology, Nanjing 210044, China (e-mail: zyy020621@163.com; yaowei@nuist.edu.cn).}
\thanks{Zhichao~Zhang is with the School of Mathematics and Statistics, Nanjing University of Information Science and Technology, Nanjing 210044, China, with the Hubei Key Laboratory of Applied Mathematics, Hubei University, Wuhan 430062, China, and also with the Key Laboratory of System Control and Information Processing, Ministry of Education, Shanghai Jiao Tong University, Shanghai 200240, China (e-mail: zzc910731@163.com).}}

\maketitle

\begin{abstract}
Graph signal processing has become an essential tool for analyzing data structured on irregular domains. While conventional graph shift operators (GSOs) are effective for certain tasks, they inherently lack flexibility in modeling dependencies between non-adjacent nodes, limiting their ability to represent complex graph structures. To address this limitation, this paper proposes the unified extended matrix (UEM) framework, which integrates the extended-adjacency matrix and the unified graph representation matrix through parametric design, so as to be able to flexibly adapt to different graph structures and reveal more graph signal information. Theoretical analysis of the UEM is conducted, demonstrating positive semi-definiteness and eigenvalue monotonicity under specific conditions. Then, we propose graph Fourier transform based on UEM (UEM-GFT), which can adaptively tune spectral properties to enhance signal processing performance. Experimental results on synthetic and real-world datasets demonstrate that the UEM-GFT outperforms existing GSO-based methods in anomaly detection tasks, achieving superior performance across varying network topologies.
\end{abstract}

\begin{IEEEkeywords}
Graph shift operator, unified extended matrix, graph signal processing, graph Fourier transform based on unified extended matrix.
\end{IEEEkeywords}

\section{Introduction}
\IEEEPARstart{G}{raph} signal processing (GSP) has emerged as a powerful framework for analyzing data structured on irregular domains \cite{ref1}–\cite{ref5}, with applications spanning sensor networks, social systems, transportation networks and biological data analysis. The core of GSP theory centers on the graph shift operator (GSO), a fundamental linear transformation that operates on the vector space of graphs signals. Once the GSO is determined, we can systematically establish the corresponding signal processing framework, thereby enabling key processing tasks for graph signals such as graph transform \cite{ref6}–\cite{ref11}, frequency analysis \cite{ref12}, \cite{ref13}, filtering \cite{ref14}–\cite{ref18}, sampling \cite{ref19}–\cite{ref24} and wavelets \cite{ref25}–\cite{ref27}.

In the GSP framework, network entities are modeled as nodes in a graph, and most GSP tools operate on a GSO matrix, which encodes pairwise relationships among nodes. The two fundamental frameworks exist for GSP: the spectral approach \cite{ref1} and the algebraic approach \cite{ref51}, each with distinct mathematical formulations and applications. The first framework utilizes the Laplacian matrix $ \mathbf{L}$. The graph Fourier transform (GFT) in this context expands a signal onto the eigenvectors of matrix $ \mathbf{L}$, with the spectrum represented by the corresponding eigenvalues. Since matrix $\mathbf{L}$ is symmetric and positive semidefinite, this approach guarantees real, non-negative eigenvalues, facilitating intuitive frequency interpretations akin to classical Fourier analysis. However, this framework is inherently limited to undirected graphs, as asymmetry in directed graphs would violate the Laplacian’s symmetry requirement. The second framework employs the adjacency matrix $ \mathbf{A}$. Here, the GFT is defined by expanding the signal into the eigenvectors of matrix $\mathbf{A}$, with the spectrum determined by its eigenvalues. Unlike the Laplacian-based approach, this method imposes no symmetry constraints on matrix $\mathbf{A}$, making it applicable to arbitrary graphs.

Alternative GFT definitions have also been proposed, such as those based on the degree matrix $\mathbf{D}$, the normalized Laplacian \cite{ref28}, or the signless Laplacian $\mathbf{Q}$ \cite{ref29}, \cite{ref30}. Fundamentally, the definition of GFT hinges on the decomposition of a general GSO \cite{ref49}, \cite{ref50}. One can define any suitable GSO tailored to their specific applications and data. Building on this concept, Averty et al. \cite{ref31} introduced a family of graph representation matrices that not only capture graph structural information more effectively but also extend previous work.

Diffusion maps (DM) \cite{ref32}–\cite{ref36} is a nonlinear dimensionality reduction method based on Markov processes, whose core concept lies in constructing a Markov matrix to reveal the intrinsic geometric structure of high-dimensional data. Unlike GSP methods, DM does not rely on a fixed GSO, but instead characterizes the geometric relationships between data points through the multiscale properties of diffusion processes.

On the other hand, in the DM framework, data points are modeled as nodes in a graph, with local similarities between nodes defined by kernel functions, e.g., Gaussian kernel \cite{ref37}. The Markov matrix transforms the Euclidean distances between high-dimensional data points into state transition probabilities, thereby describing the random walk process between data states. Through eigen-decomposition of this matrix, its eigenvectors and eigenvalues can be obtained to construct the diffusion distances (DDs). The DDs considers all possible transition paths, effectively characterizing the global connectivity between data points and providing a novel perspective for analyzing complex data structures.
Notably, in this framework, each node corresponds to a complete data state, while edge weights reflect the transition relationships between these high-dimensional states. This modeling approach contrasts sharply with GSP methods, which view edges as local connections between individual network elements. 

Then, Heimowitz et al. \cite{ref38} proposed using the Markov matrix as a GSO, establishing a connection between DM and GSP. Research demonstrates that the Markov matrix not only possesses ideal mathematical properties such as diagonalizability but also enables efficient computation of the inverse eigenvector matrix. More importantly, when the Markov matrix is employed as the GSO, the GSP framework exhibits profound correlations with the DM framework, laying a theoretical foundation for the integrated application of these two methodologies.

Furthermore, Elias et al. \cite{ref39} integrated DDs into GSP, developing a graph model that captures interactions in Markov networks. While previous work directly employs a non-symmetric Markov matrix as the GSO, the present study adopts a fundamentally different approach by using a doubly-stochastic matrix derived from discrete-time consensus algorithms \cite{ref40} as the initial basis for GSP. The resulting extended adjacency matrix GSO exhibits symmetry and captures dependencies between non-adjacent nodes, with its connection patterns dynamically adapting to changes in the diffusion scale. Similar to the extended-adjacency matrix, the corresponding Laplacian matrix and degree matrix can be defined. This raises a legitimate question: which matrix proves most effectively? The optimal choice remains an open question, as the selection depends critically on both the target application domain, the specific graph operations required and data.

\begin{figure}[!t]
	\centering
	\includegraphics[width=3.0in]{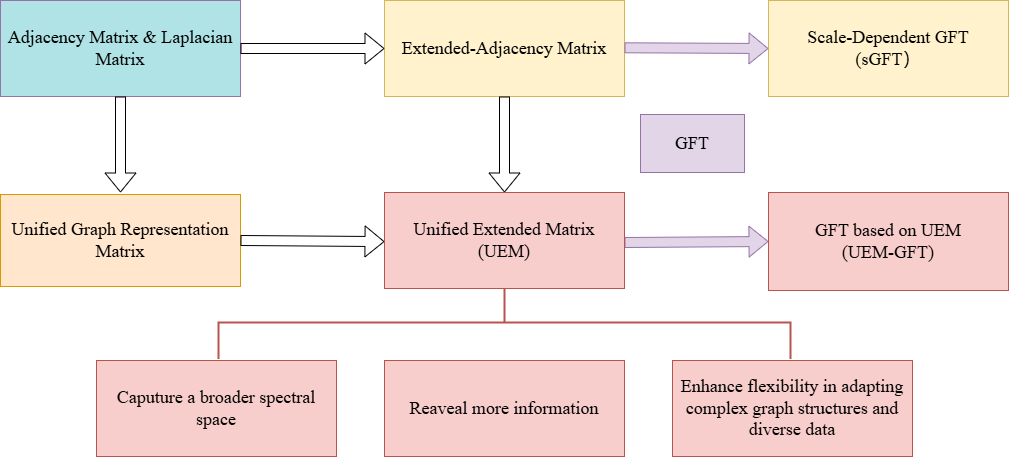}
    \caption{A novel parametric framework based on extended-adjacency matrix.}
    \label{fig_9}
\end{figure}

Inspired by unified graph representation matrix, we extend the work of Elias et al. The new framework can be seen in Fig. \ref{fig_9}. In this paper, we propose unified extended matrix (UEM) and discuss its application. The main contributions of this paper are as follows:
\begin{itemize}
\item { The UEM is introduced as a novel framework. By enabling parameter adjustment of the UEM to capture a broader spectral space and achieve more flexible spectral characteristics, it enhances adaptability to complex graph structures and diverse data, overcoming the limitations of existing frameworks.}
\item { We derive two key theoretical properties of the UEM, with the second one further confirmed via simulation example on synthetic graph.}
\item { We propose the GFT based on UEM (UEM-GFT) and provide comprehensive experiments on synthetic and real-world datasets, with detailed comparisons with different approaches to demonstrate the benefits of UEM-GFT.}
\end{itemize}

The rest of the manuscript is organized as follows: Section II reviews some relevant concepts of GSP and two GSOs. In section III, we present the theoretical foundations of UEM-GFT and present simulation example that illustrates the proposed property. In Section IV, we consider the application of UEN-GFT in anomaly detection and, finally, in Section V, we draw a conclusion.

\section{Graph Signal Processing (GSP)} 
In this section, we first briefly review concepts of GSP that are relevant to this paper. Next, we introduce two GSOs, those are unified graph representation matrix and extended-adjacency matrix.
\subsection{Graph Signals}
Let a graph be represented by $\cal G = \{\mathcal{V},\mathcal{E},\mathbf{A}\}$, where $\mathcal{V} =\{v_1, \dots, v_N\}$ denotes the set of nodes with the graph, $\mathcal{E} = \{e_{11}, \dots, e_{NN}\}$ denotes the set of edges and adjacency matrix \( \mathbf{A} \) is the weighted adjacency matrix. The adjacency matrix $\mathbf{A} \in \mathbb{C}^{N \times N}$ denotes the edge weights between the nodes of the graph, whose $(i,j)^{\text{th}}$ element is $\mathbf{A}_{ij}=w_{ij},\forall\, i,j \in \{1,\dots,N\}$. And weights $w_{ij}$ can be added to edge to signify importance of a link between two nodes in a network. 

 If the graph is undirected, the relation goes both ways, $\mathbf{A}_{ij}=\mathbf{A}_{ji}$, and the nodes are neighbors. In the unweighted case, matrix \( \mathbf{A} \) is binary, where the coefficient $\mathbf{A}_{ij}$ is 1 if there exists an edge $\{i, j\}$ and 0 otherwise. For weighted graph, the 1 is replaced by the weight $w_{ij}$. The graph Laplacian $\mathbf{L} = \mathbf{D} - \mathbf{A}$ constitutes another prevalent matrix, with the diagonal degree matrix $\mathbf{D}$ defined by $\deg(i) = \sum_{j=1}^n \mathbf{A}_{ij}$ for node connectivity information. 

A graph signal is a discrete function $x: \mathcal{V} \to \mathbb{C}$ that assigns a complex-valued scalar $\mathbf{x}_n \in \mathbb{C}$ to each vertex $\mathbf{v}_n \in \mathcal{V}$. Equivalently, the signal can be represented as a complex-valued column vector $\mathbf{x} = [x_1, x_2, \dots, x_N]^T$, $\mathbf{x} \in \mathbb{C}^N$, where the $n^{\text{th}}$ component $\mathbf{x}_n$ corresponds to the signal value at vertex $\mathbf{v}_n$. 

\subsection{Graph Shift Operator (GSO)}
The structure of a graph $\mathcal{G}$ is represented by a matrix $\mathbf{S} \in \mathbb{R}^{N \times N}$, referred to as the GSO. For   to qualify $\mathbf{S}$ as a valid GSO, its entries must satisfy the sparsity constraint:
\begin{equation}
	[\mathbf{S}]_{ji} = s_{ji} = 0 \quad \text{whenever} \ (i, j) \notin \mathcal{E} \ \text{for} \ i \neq j.
\end{equation}

Different choices of $\mathbf{S}$ exist depending on the application context. Common variants include the adjacency matrix $\mathbf{A}$ and the Laplacian matrix $\mathbf{L}$, both of which are specific instances of $\mathbf{S}$. The adjacency matrix $\mathbf{A}$ is often selected due to analogy with discrete-time shift operations in classical signal processing, making it particularly suitable for modeling direct neighbor, e.g., in social network propagation modeling. In contrast, the Laplacian matrix $\mathbf{L}$ is preferred in settings requiring spectral analysis, as it facilitates graph frequency-domain interpretations rooted in spectral graph theory. Let the adjacency matrix $\mathbf{A}$ and the Laplacian matrix $\mathbf{L}$ of a graph $\cal G$ have ordered eigenvalues $\lambda_1 \leq \cdots \leq \lambda_n$ and $0 = \mu_1 \leq \cdots \leq \mu_n$, respectively. These spectral characteristics encode important structural information about the graph, enabling the derivation of key topological properties.

\subsection{Graph Fourier Transform (GFT)}
In the definition of the GFT, the GSO S is assumed to be diagonalizable. Consider the eigen-decomposition of $\mathbf{S}$:
\begin{equation}
	\mathbf{S} = \mathbf{U}\mathbf{\Lambda}\mathbf{U}^{-1},
\end{equation}
where $\mathbf{U} = [\mathbf{u}_1, \mathbf{u}_2, \dots, \mathbf{u}_N]$ is composed of the eigenvectors as column vectors and $\mathbf{\Lambda} = \text{diag}(\boldsymbol{\lambda})$ is a diagonal matrix with the corresponding eigenvalues $\boldsymbol{\lambda} = [\lambda_1, \dots, \lambda_N]$. 
Let $\mathbf{x}$ be a graph signal on graph $\cal G$ with matrix $\mathbf{S}$. The GFT matrix is then defined as $\mathbf{F}\ = \mathbf{U}^{-1}$   such that the GFT of $\mathbf{x}$ is given by
\begin{equation}
	\hat{\mathbf{x}} = \mathbf{F} \mathbf{x} = \mathbf{U}^{-1} \mathbf{x},
\end{equation}
where $\hat{\mathbf{x}}$ denotes the graph signal in the graph Fourier domain. And the inverse transformation to the vertex domain can be given by
\begin{equation}
	\mathbf{x} = \mathbf{F}^{-1}\hat{\mathbf{x}} = {\mathbf{U}}\hat{\mathbf{x}}.
\end{equation}

\subsection{Unified Graph Representation Matrix}
The adjacency matrix $\mathbf{A}$ and the Laplacian matrix $\mathbf{L}$ are the two main graph representations but they are singular. Therefore, Averty et al. proposed the unified graph representation matrix which contains the classical representation matrices building on previous work. The unified graph representation matrix is constructed:
\begin{equation}
	\mathbf{P}_{m,n} := m\mathbf{D} + (2n - 1)(m - 1)\mathbf{A}, \quad m,n \in [0, 1].
\end{equation}
The matrix $\mathbf{P}_{m,n}$  with different values of $m$ and $n$ degenerates into the classical matrix. Obviously,
\begin{equation}
\mathbf{A}=\mathbf{P}_{0.0,0.0}, \mathbf{L}=2\mathbf{P}_{0.5,1.0}, \mathbf{D}=\mathbf{P}_{1.0,n}.
\end{equation}
\subsection{Extended-Adjacency Matrix}
While conventional adjacency matrix $\mathbf{A}$ and the Laplacian matrix $\mathbf{L}$  only capture relationships between directly connected nodes, Elias et al. proposed the extended-adjacency matrix $\bar{\mathbf{A}}(t)$, which additionally accounts for dependencies between non-adjacent nodes. Let the network be represented by a connected graph $\cal G = \{\mathcal{V},\mathbf{B}\}$, where the adjacency matrix $\mathbf{B}$ is symmetric, irreducible, stochastic, and contains only positive real entries for existing edges. This matrix is equivalent to Markov matrix, where each node in the graph corresponds to a distinct state in the Markov chain. The diffusion distance between nodes $v_i$ and $v_j$ in a graph is expressed as:
\begin{equation}
	D_t^2(v_i, v_j) = \sum_{n=1}^N \frac{\left( B_{in}^{(t)} - B_{jn}^{(t)} \right)^2}{1/N},
\end{equation}
where $B_{ij}^{(t)}$ denotes the $(i, j)^{th}$ entry of $\mathbf{B}^{t}$, representing the probability of reaching nodes $v_j$ from nodes $v_i$ after $t$ steps of random walk. It is used in the calculation of diffusion distance to quantify the similarity between nodes.

And the extended-adjacency matrix $\bar{\mathbf{A}}(t)$ is such that
\begin{equation}
	\bar{A}_{ij}(t) = 
	\begin{cases}
		{B}_{ij} + \exp\left(-\dfrac{D_t^2(v_i, v_j)}{\rho N}\right) & \text{ } i \neq j \\
		0 & \text{ } i = j,\label{eq:adjacency_nonneg}
	\end{cases}
\end{equation}
where $\rho$ is a free parameter and $N$ is the size of the network. In Eq. \eqref{eq:adjacency_nonneg}, the component $B_{ij}$ preserves the original graph edges, while the RBF term facilitates adjacency extension. Notably, the ${\rho N}$ scaling factor ensures network-size invariance in the RBF kernel argument, with the output range of kernel being application-adjustable through parameter $\rho$.
\subsection{Scale-Dependent GFT (sGFT)}
Based on the extended-adjacency matrix $\bar{\mathbf{A}}(t)$, whose entries depend on the diffusion scale $t$, the relevant matrices can be defined. For each scale $t$, the extended-Laplacian matrix $\bar{\mathbf{L}}(t)$ is given by:
\begin{equation}
	\bar{\mathbf{L}}(t) = \bar{\mathbf{D}}(t)-\bar{\mathbf{A}}(t),
\end{equation}
where $\bar{\mathbf{D}}(t)$ is the associated diagonal degree matrix. Using the eigen-decomposition $\bar{\mathbf{L}}(t) =\bar{\mathbf{U}}(t) \bar{\mathbf{\Lambda}}(t) \bar{\mathbf{U}}^{-1}(t)$, the sGFT of a signal $\mathbf{x}$ is:
\begin{equation}
	\hat{\mathbf{x}}(t) = \bar{\mathbf{U}}^{-1}(t) \mathbf{x}.
\end{equation}
Unlike the conventional GFT with fixed graph-frequency coefficients, the coefficients $\hat{\mathbf{x}}(t)$ vary with the diffusion scale $t$. The corresponding scale-dependent graph Fourier synthesis equation is:
\begin{equation}
	\mathbf{x} = \bar{\mathbf{U}}(t)\hat{\mathbf{x}}(t).
\end{equation}

\section{Unified Extended Matrix (UEM)}
In this section, we formally introduce the definition of UEM and prove two properties. To validate the second property and the effect of UEM in capturing more information, we conduct simulation examples on synthetic graphs.
\subsection{Definition}
The extended-adjacency matrix has limited flexibility in capturing dependencies between non-adjacent nodes, restricting its effectiveness for modeling complex graph topologies. To overcome this, we propose the UEM framework, which integrates the extended-adjacency matrix with the unified graph representation matrix via a parameterized scheme. This design enables UEM to adapt flexibly to complex graph structures and reveal more graph signal information. The UEM is defined as:
\begin{equation}
	\bar{\mathbf{P}}_{m,n}(t) := m\bar{\mathbf{D}}(t) + (2n - 1)(m - 1)\bar{\mathbf{A}}(t),
\end{equation}
where $m,n \in [0, 1]$ and $t \in \mathbb{N}$.

Comparisons with the unified graph representation matrix: The UEM framework extends the unified graph representation matrix $\mathbf{P}_{m,n}$ by substituting the traditional adjacency matrix and degree matrix with their extended counterparts: the extended-adjacency matrix and the extended-degree matrix, respectively. This extension preserves the parametric structure of $\mathbf{P}_{m,n}$ while incorporating the diffusion scale $t$. Crucially, whereas $\mathbf{P}_{m,n}$ relies solely on local adjacency information, $\bar{\mathbf{P}}_{m,n}(t)$ integrates dependencies between non-adjacent nodes via diffusion distances.

Comparisons with extended-adjacency and extended-Laplacian matrices: The UEM $\bar{\mathbf{P}}_{m,n}(t)$ shares the diffusion scale $t$ with extended-adjacency matrix $\bar{\mathbf{A}}(t)$ and extended-Laplacian matrix $\bar{\mathbf{L}}(t)$ to capture non-adjacent node dependencies, overcoming the limitation of traditional matrices that only reflect direct adjacencies. And $\bar{\mathbf{P}}_{m,n}(t)$ can degenerate to these specific matrices by adjusting parameters $m$ and $n$:
\begin{equation}
	\bar{\mathbf{A}}(t) = \bar{\mathbf{P}}_{0.0,0.0}(t), 
	\bar{\mathbf{L}}(t) = 2\bar{\mathbf{P}}_{0.5,1.0}(t),
	\bar{\mathbf{D}}(t) = \bar{\mathbf{P}}_{1.0,n}(t).
\end{equation}
Crucially, unlike $\bar{\mathbf{A}}(t)$ and $\bar{\mathbf{L}}(t)$, which depend only on the diffusion scale $t$, $\bar{\mathbf{P}}_{m,n}(t)$ incorporates the parameters $m$, $n$ and $t$. This multiparameter formulation enables more flexible modeling of complex graph structures.
\subsection{Properties}
The resulting UEM $\bar{\mathbf{P}}_{m,n}(t)$ is a symmetric matrix that depend on the parameters $m$, $n$ and diffusion scale $t$. In the following, we present two important properties of $\bar{\mathbf{P}}_{m,n}(t)$, namely the positive semi-definiteness and the monotonicity of their eigenvalues in $m$.

\textit{Proposition 1:} Let $\mathcal{G}$ be a graph with its representation plan. Then $\bar{\mathbf{P}}_{m,n}(t)$ is positive semidefinite if 
\[
\frac{2m - 1}{2(m - 1)} \leq n \leq \frac{1}{2(1 - m)}.
\]

\textit{Proof:} Since \(\bar{\mathbf{A}}(t) = \bar{\mathbf{D}}(t) - \bar{\mathbf{L}}(t)\), the expression of \(\bar{\mathbf{P}}_{m,n}(t)\) can be written as:
\begin{equation}
	\bar{\mathbf{P}}_{m,n}(t) = \bigl[m + (2n-1)(m-1)\bigr]\bar{\mathbf{D}}(t) + (2n-1)(1-m)\bar{\mathbf{L}}(t).
\end{equation}
	From Eq. \eqref{eq:adjacency_nonneg}, for \(\forall i, j\), \(\bar{A}_{ij}(t) \geq 0\). The quadratic form of \(\bar{\mathbf{L}}(t)\) satisfies:
\begin{equation}
	\begin{split}
		\langle \bar{\mathbf{L}}(t)\mathbf{x}, \mathbf{x} \rangle &= \langle \bar{\mathbf{D}}(t)\mathbf{x}, \mathbf{x} \rangle - \langle \bar{\mathbf{A}}(t)\mathbf{x}, \mathbf{x} \rangle \\
		&= \sum_{(i,j) \in \mathcal{E}} \bar{A}_{ij}(t)(x_i - x_j)^2 \geq 0,
	\end{split}\label{eq:Laplacian_nonneg}
\end{equation}
where
\begin{equation}
	\begin{aligned}
		\langle \bar{\mathbf{D}}(t)\mathbf{x}, \mathbf{x} \rangle &= \sum_{i=1}^{N} \left( \sum_{j=1}^{N} \bar{A}_{ij}(t) \right) x_i^2 \\
		&= \sum_{(i,j) \in \mathcal{E}} \bar{A}_{ij}(t)(x_i^2 + x_j^2) \geq 0,
	\end{aligned}\label{eq:degree_nonneg}
\end{equation}
and
\begin{equation}
	\begin{split}
		\langle \bar{\mathbf{A}}(t)\mathbf{x}, \mathbf{x} \rangle &= \sum_{i=1}^{N} \sum_{j=1}^{N} \bar{A}_{ij}(t)x_i x_j = 2 \sum_{(i,j) \in \mathcal{E}} \bar{A}_{ij}(t)x_i x_j.
	\end{split}
\end{equation}
From Eqs. \eqref{eq:Laplacian_nonneg} and \eqref{eq:degree_nonneg}, $\bar{\mathbf{L}}(t)$ and $\bar{\mathbf{D}}(t)$ are two positive semidefinite matrices. To ensure the positive semi-definiteness of $\bar{\mathbf{P}}_{m,n}(t)$, the condition $ m + (2n - 1)(m - 1) \geq 0 $ and  $(2n - 1)(1- m) \geq 0 $  must be satisfied. Thus, it implies $ \frac{1}{2} \leq n \leq \frac{1}{2(1-m)} $. Secondly, generalizing what Averty et al did for $\mathbf{P}_{m,n}$ to $\bar{\mathbf{P}}_{m,n}(t)$: 
\begin{equation}
	\begin{aligned}
		\left\langle \bar{\mathbf{P}}_{m,n}(t)\mathbf{x}, \mathbf{x} \right\rangle 
		&=\Big[ m - (2n-1)(m-1) \Big]\sum_{(i,j)\in \mathcal{E}} \bar{A}_{ij}(t)x_i^2 \\
		&\quad+\Big[ m - (2n-1)(m-1) \Big]\sum_{(i,j)\in \mathcal{E}} \bar{A}_{ij}(t) x_j^2 \\
		&\quad + (2n-1)(m-1) \sum_{(i,j)\in \mathcal{E}} \bar{A}_{ij}(t)(x_i + x_j)^2,
	\end{aligned}
\end{equation}
where \( x_i \) is the \( i^{th} \) component of the vector \( \mathbf{x} \in \mathbb{R}^N \). Then, another condition needed to guarantee the positive semi-definiteness is to have $ m - (2n - 1)(m - 1) \geq 0 $ and $(2n-1)(m-1) \geq 0$, which implies $\frac{2m - 1}{2(m - 1)} \leq n \leq \frac{1}{2}$. Finally, by grouping all the conditions together, we obtain a sufficient condition: if  $\frac{2m - 1}{2(m - 1)} \leq n \leq \frac{1}{2(1-m)}$, then $\bar{\mathbf{P}}_{m,n}(t)$ is positive semidefinite.

Let us denote $v^{(m,n)}_{l}(t)$ the eigenvalue of the matrix $\bar{\mathbf{P}}_{m,n}(t)$. Another interesting result is the generalization of Avertv's Proposition 2 on the monotonicity of $v^{(m,n)}_{l}$.

\textit{Proposition 2:} Let $\mathcal{G}$ be a graph, $m \in [0,1]$ and $m' \in [m,1]$. Then,
\begin{equation}
	v^{(m,n)}_{l}(t) \leq v^{(m',n)}_{l}(t), \quad \forall n \in [0,1], \, t \in \mathbb{N}.
\end{equation}

\textit{Proof:} Let $m \in [0,1]$ and $m' \in [m,1]$. The same proof scheme as Avertv is used, so the following expression needs to be written:
	\begin{equation}
		\bar{\mathbf{P}}_{m',n}(t) - \bar{\mathbf{P}}_{m,n}(t) = (m'-m)\bigl[\bar{\mathbf{D}}(t)-(2n-1)\bar{\mathbf{A}}(t)\bigr].
	\end{equation}
	Thanks to a simplified form of the Weyl theorem \cite{ref41}, it follows:
	\begin{equation}
		v_{l}\bigl[\bar{\mathbf{P}}_{m',n}(t)\bigr] - v_{l}\bigl[\bar{\mathbf{P}}_{m,n}(t)\bigr] \geq (m'-m)v_{1}\bigl[\bar{\mathbf{D}}(t)-(2n-1)\bar{\mathbf{A}}(t)\bigr],
	\end{equation}
where \( v_1 \left[\bar{\mathbf{D}}(t) - (2n-1)\bar{\mathbf{A}}(t) \right] \), which denotes the smallest eigenvalue of the matrix $\bar{\mathbf{M}}(t) :=\bar{ \mathbf{D}}(t) - (2n-1)\bar{\mathbf{A}}(t)$ is non-negative. To prove it, let us show that $\bar{\mathbf{M}}(t)$  is positive semidefinite. Let \(\mathbf{x}\) be a vector in \(\mathbb{R}^N\):
\begin{equation}
	\begin{aligned}
		\langle \bar{\mathbf{M}}(t)\mathbf{x}, \mathbf{x} \rangle 
		&= \sum_{(i,j) \in \mathcal{E}} \bar{A}_{ij}(t)\bigl(x_i^2 + x_j^2\bigr) \\
		&\quad - 2(2n-1) \sum_{(i,j) \in \mathcal{E}} \bar{A}_{ij}(t)x_i x_j \\
		&= \sum_{(i,j) \in \mathcal{E}} \bar{A}_{ij}(t)
		\bigl[x_i^2 + x_j^2 - 2(2n-1)x_i x_j\bigr] \geq 0.
	\end{aligned}
\end{equation}
This implies $\bar{\mathbf{M}}(t)$ has no negative eigenvalues. The non-negativity of the smallest eigenvalue in particular establishes the result, concluding the proof.

\subsection{GFT based on UEM (UEM-GFT)}
Let the diagonalizable matrix $\bar{\mathbf{P}}_{m,n}(t)$ be a GSO, with the eigen-decomposition:
\begin{equation}
	\bar{\mathbf{P}}_{m,n}(t) = \bar{\mathbf{U}}_{m,n}(t) \bar{\mathbf{\Lambda}}_{m,n}(t) \bar{\mathbf{U}}_{m,n}^{-1}(t),
\end{equation}
Then, the UEM-GFT of \( \mathbf{x} \) is given by:
\begin{equation}
	\hat{\mathbf{x}}_{m,n}(t) = \bar{\mathbf{U}}_{m,n}^{-1}(t)\mathbf{x},
\end{equation}
where \( \hat{\mathbf{x}}_{m,n}(t) \) denotes the graph signal in the graph Fourier domain, which depends on parameters \( m \), \( n \) and \( t \). The inverse transformation to the vertex domain is given by:
\begin{equation}
	\mathbf{x} = \bar{\mathbf{U}}_{m,n}(t) \hat{\mathbf{x}}_{m,n}(t).
\end{equation}

Comparisons with the GFT: The UEM-GFT, like the GFT, is based on the eigen-decomposition of GSO. However, the GFT is defined based on fixed GSOs such as the adjacency matrix $\mathbf{A}$ or Laplacian matrix $\mathbf{L}$, whose spectral characteristics  are inherently determined by the graph structure and cannot be adjusted, limiting their adaptability to complex graph structures. In contrast, the UEM-GFT employs the UEM $\bar{\mathbf{P}}_{m,n}(t)$ as its GSO, which integrates the extended-adjacency matrix and unified graph representation matrix through parametric design. This allows UEM-GFT to dynamically tune its spectral properties via parameters $m$, $n$ and $t$, enabling it to capture a broader spectral space, thereby overcoming the intrinsic limitations of the GFT.

Comparisons with the sGFT: Both the UEM-GFT and the sGFT overcoming the limitations of traditional GSOs by incorporating diffusion scale $t$ to model dependencies between non-adjacent nodes, enhancing the capability for processing complex graph signals compared to the GFT. Significantly, when $m=0.5$ and $n=1.0$, the UEM-GFT reduces to the sGFT. And the critical distinction lies in the flexibility of GSOs: the sGFT utilizes extended-Laplacian matrix $\bar{\mathbf{L}}(t)$, whose spectral properties depend exclusively on diffusion scale $t$. In contrast, the UEM-GFT introduces additional parameters $m$ and $n$ to the framework. This parametric integration allows the UEM-GFT to not only capture non-local dependencies via $t$ but also flexibly adapt to different graph structures through $m$ and $n$. 

\subsection{Simulation Examples}
\begin{figure}[!t]
	\centering
	\subfloat{\includegraphics[width=1.7in]{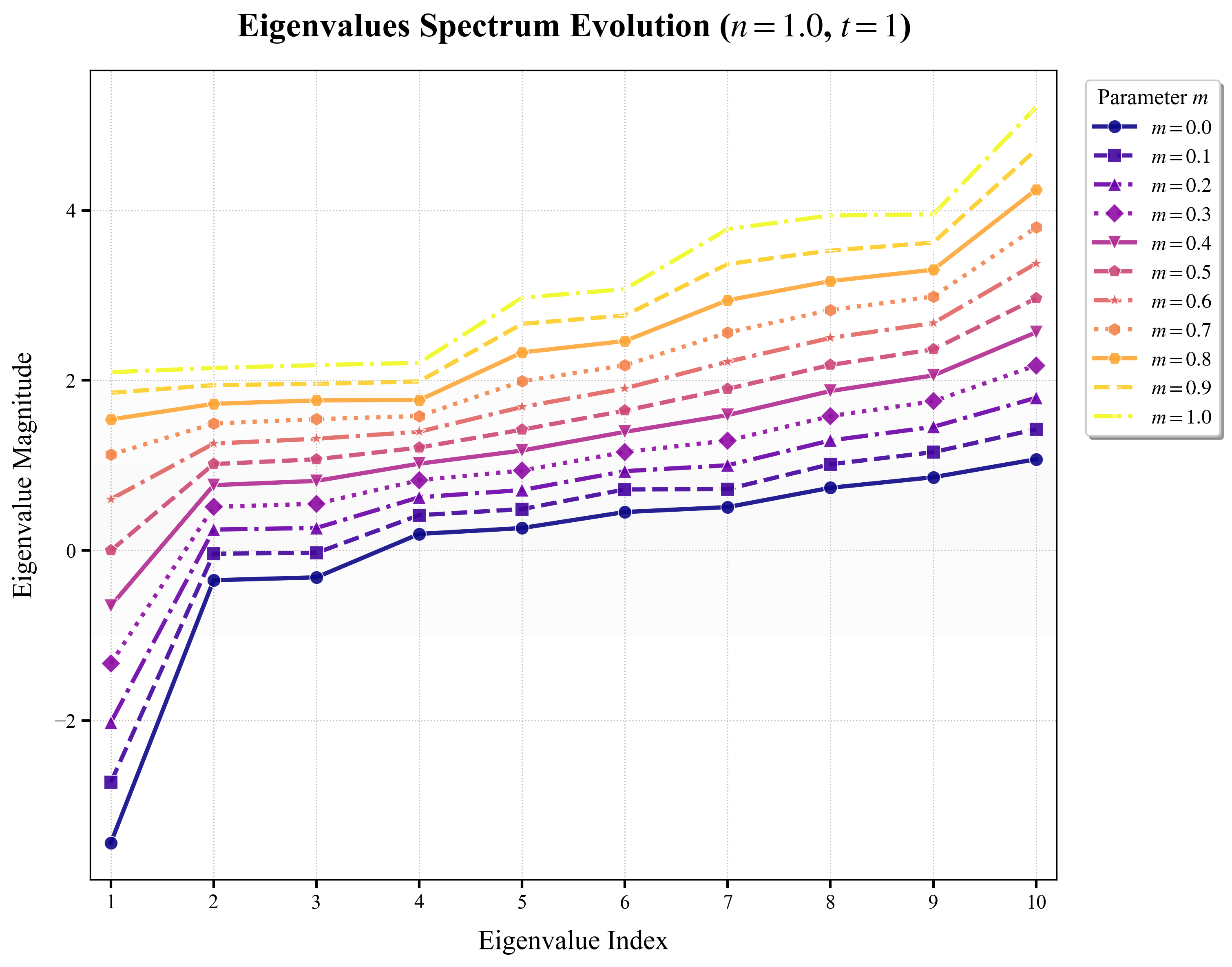}}
	\subfloat{\includegraphics[width=1.7in]{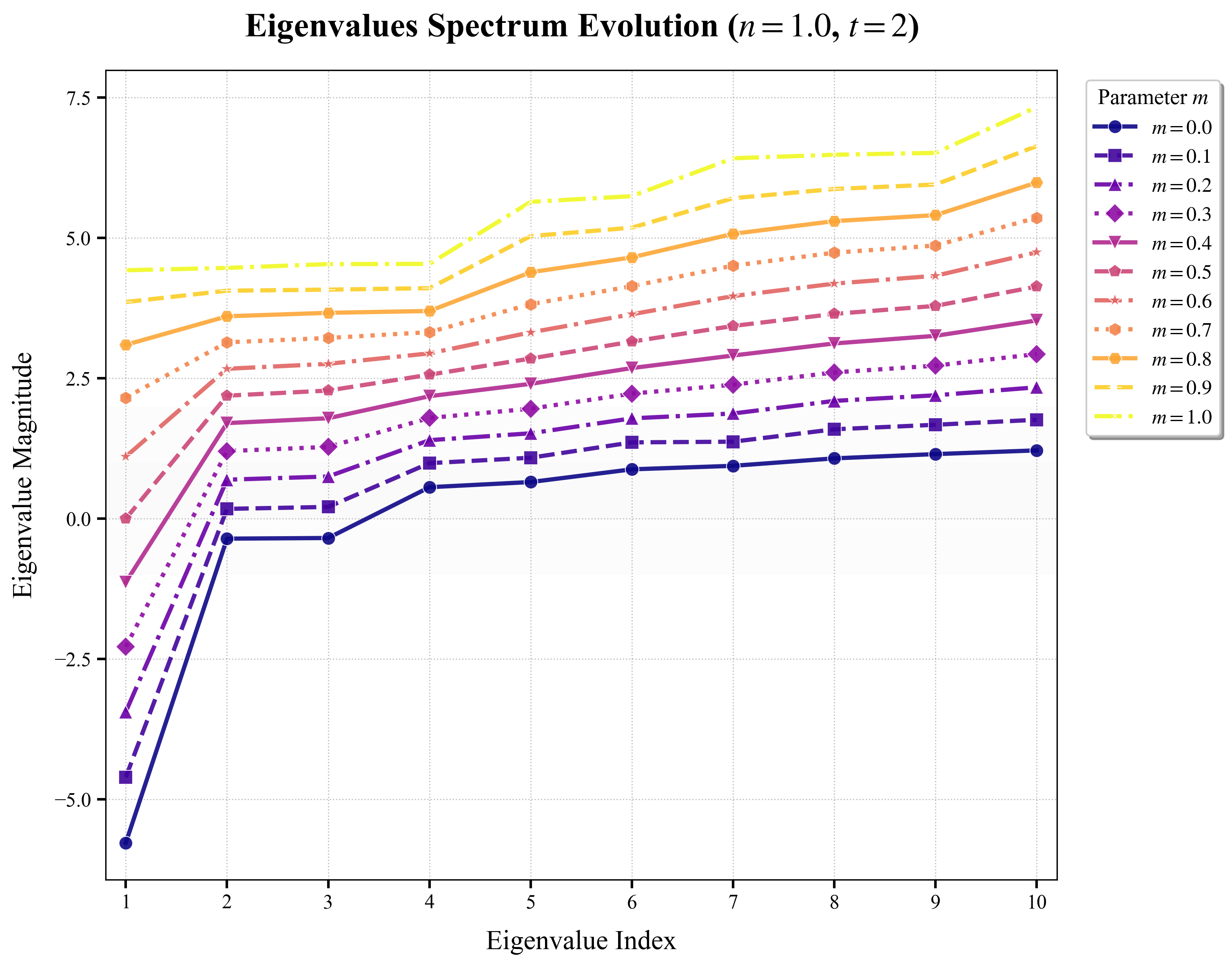}}
	\caption{Figures (Left: $t=1$/Right: $t=2$) depict the variations of eigenvalues for different parameters $m$ with $n = 1.0$.}
	\label{fig 1}
\end{figure}

In order to confirm the monotonicity of eigenvalues with respect to parameter $m$, we provide a simulation example on a sensor network graph. Consider an unweighted and undirected graph $\mathcal{G} = (V, \mathbf{A})$ which represents the network topology, where $\mathbf{L}$ denotes the Laplacian matrix. The matrix $\mathbf{Z} = \mathbf{I} - \epsilon\mathbf{L}$ was introduced in previous work, where $\epsilon$ is the consensus step size \cite{ref40}, with $\mathbf{B} = \mathbf{Z}$. In this paper, the parameter $\epsilon$ is set to $\epsilon = \frac{1}{1.25\Delta}$, where $\Delta$ is the maximum degree in $\mathcal{G}$. We consider a sensor network modeled as a $k$-nearest neighbor ($k$-NN) graph with the following parameters $N = 10$ sensors, $k = 3$ and $\rho = 0.4$. After eigen-decomposing each matrix, the resulting eigenvalues are sorted in ascending order. Fig. \ref{fig 1} shows that all eigenvalues are monotonically non-decreasing with increasing parameter $m$ when $n = 1.0$ and \(t \in \{1, 2\}\). This simulation validates the effects of parameter $m$ on eigenvalues as presented in Proposition 2.

\begin{figure}[!t]
	\centering
	\subfloat[]{\includegraphics[width=1.7in]{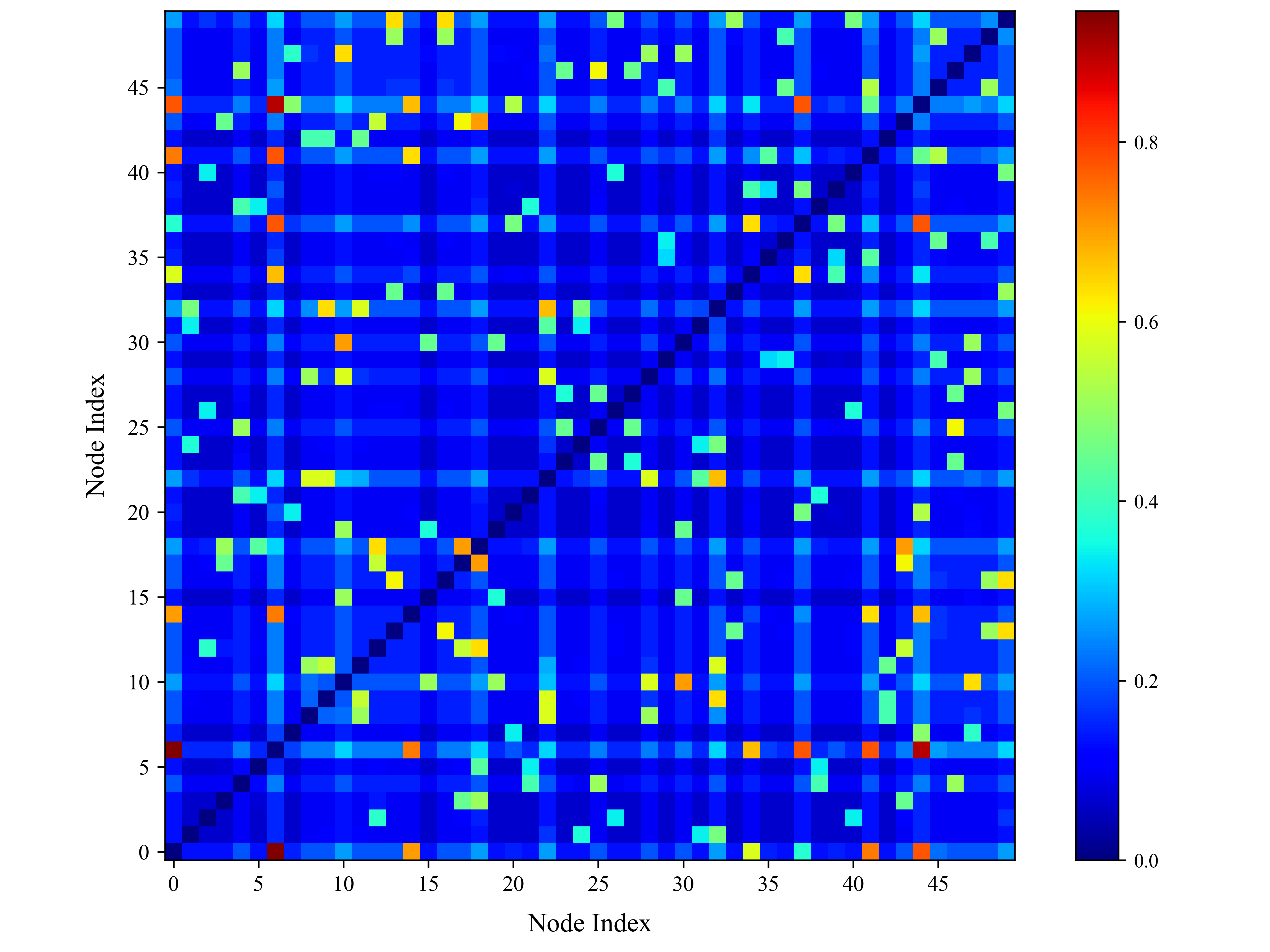}%
		\label{fig_first}}
	\hfil
	\subfloat[]{\includegraphics[width=1.7in]{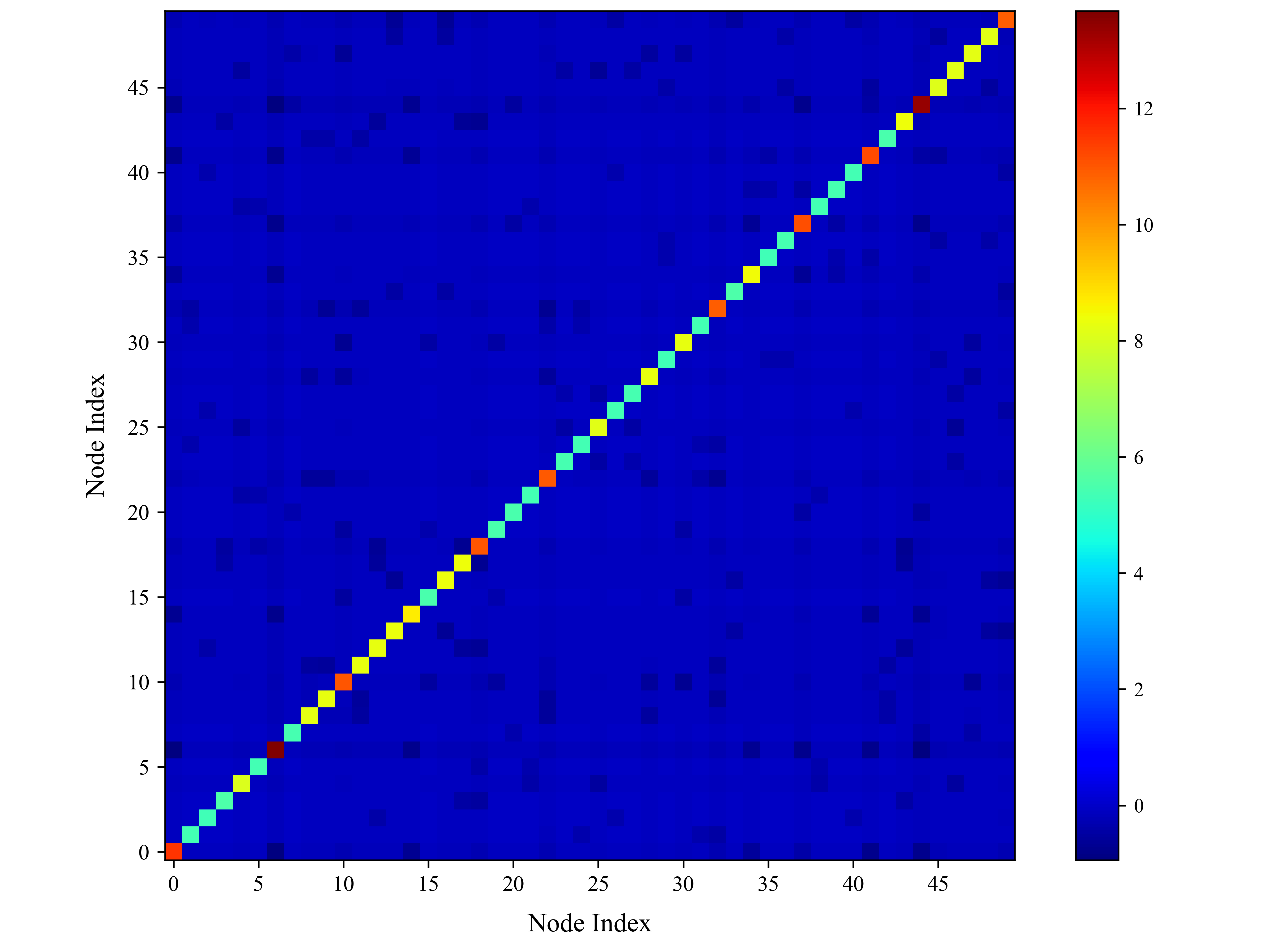}%
		\label{fig_second}}
	\hfil
	\subfloat[]{\includegraphics[width=1.7in]{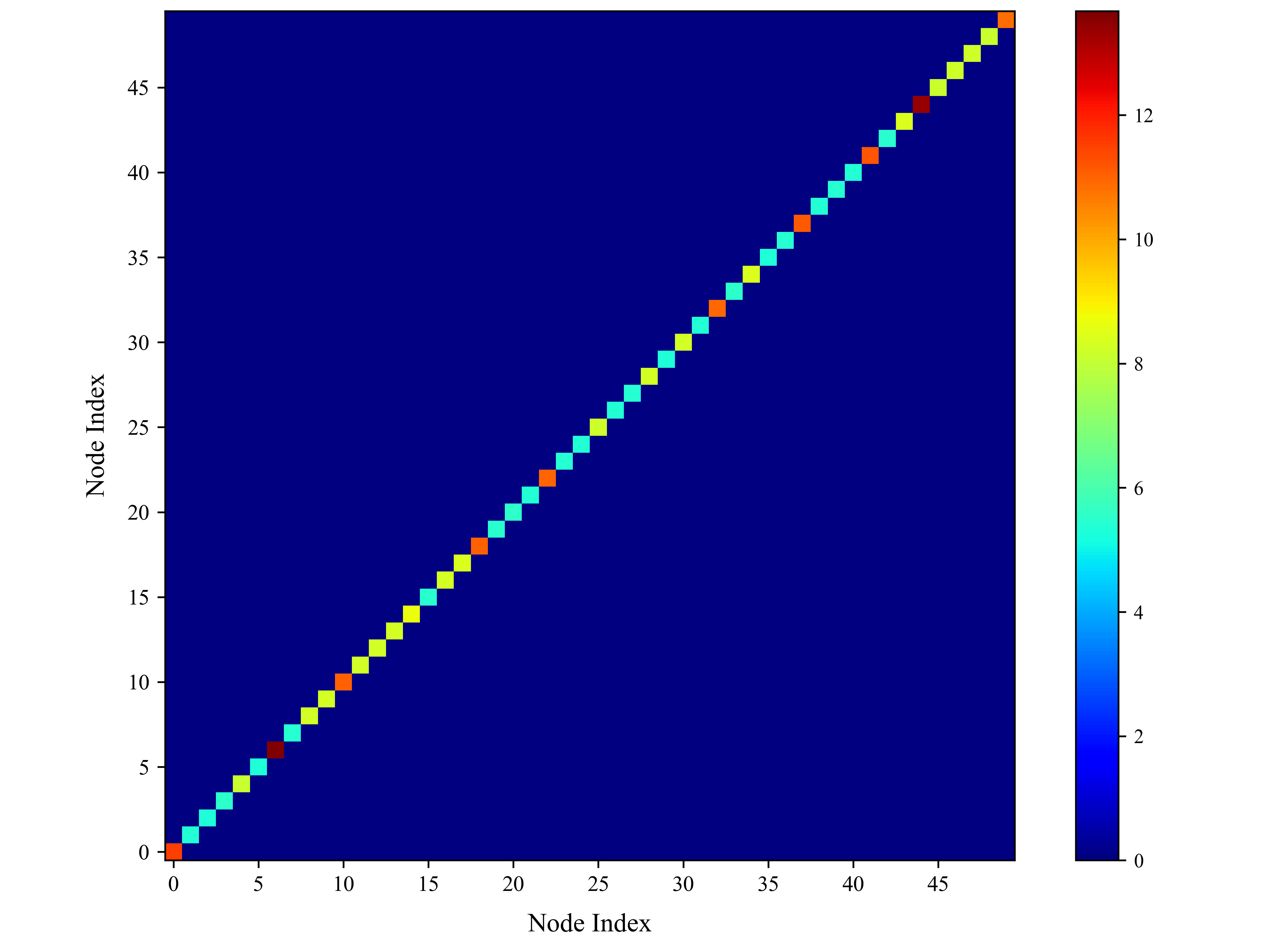}%
		\label{fig_third}}
	\hfil
	\subfloat[]{\includegraphics[width=1.7in]{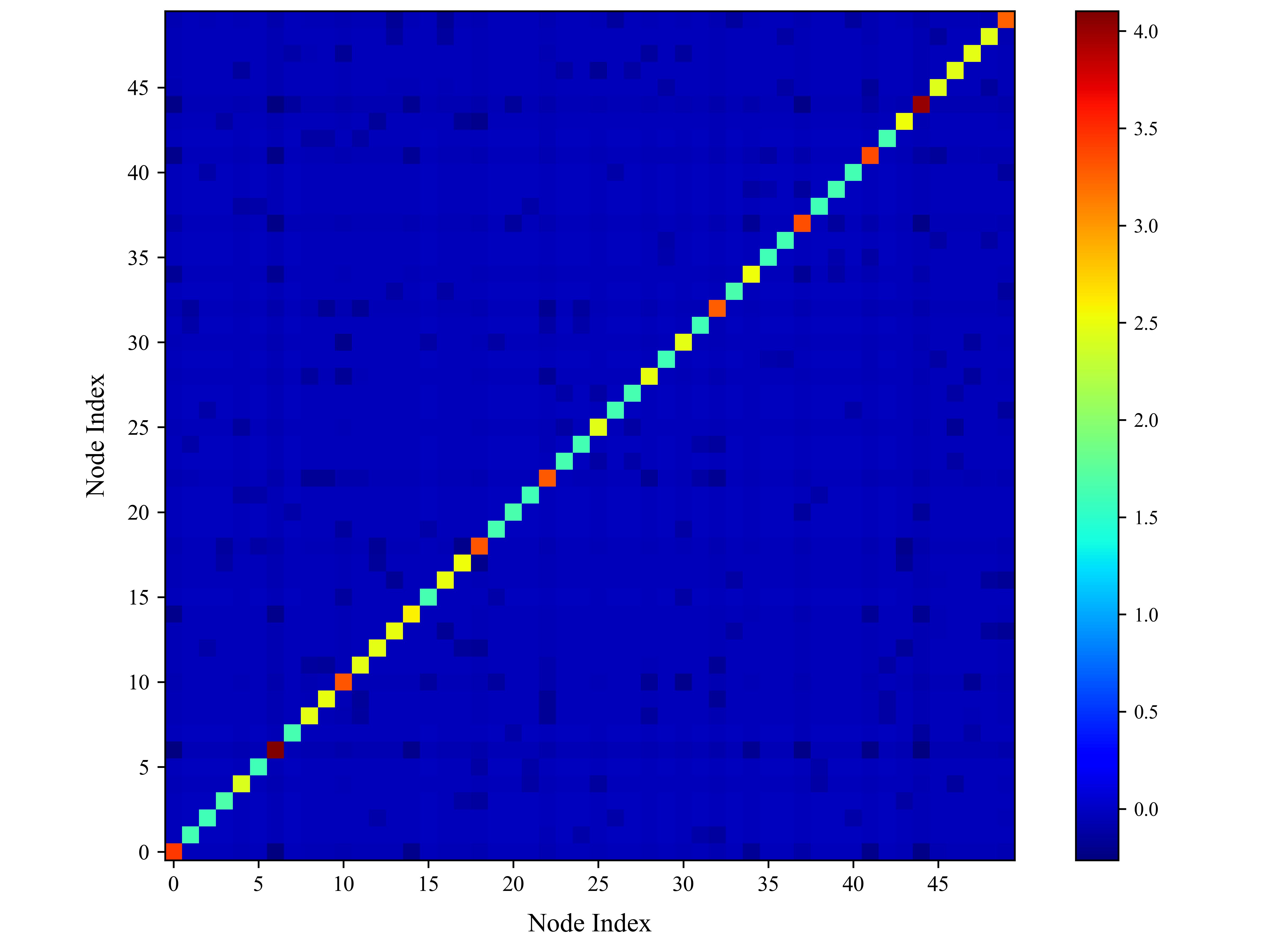}%
		\label{fig_fourth}}
	\hfil
	\subfloat[]{\includegraphics[width=1.7in]{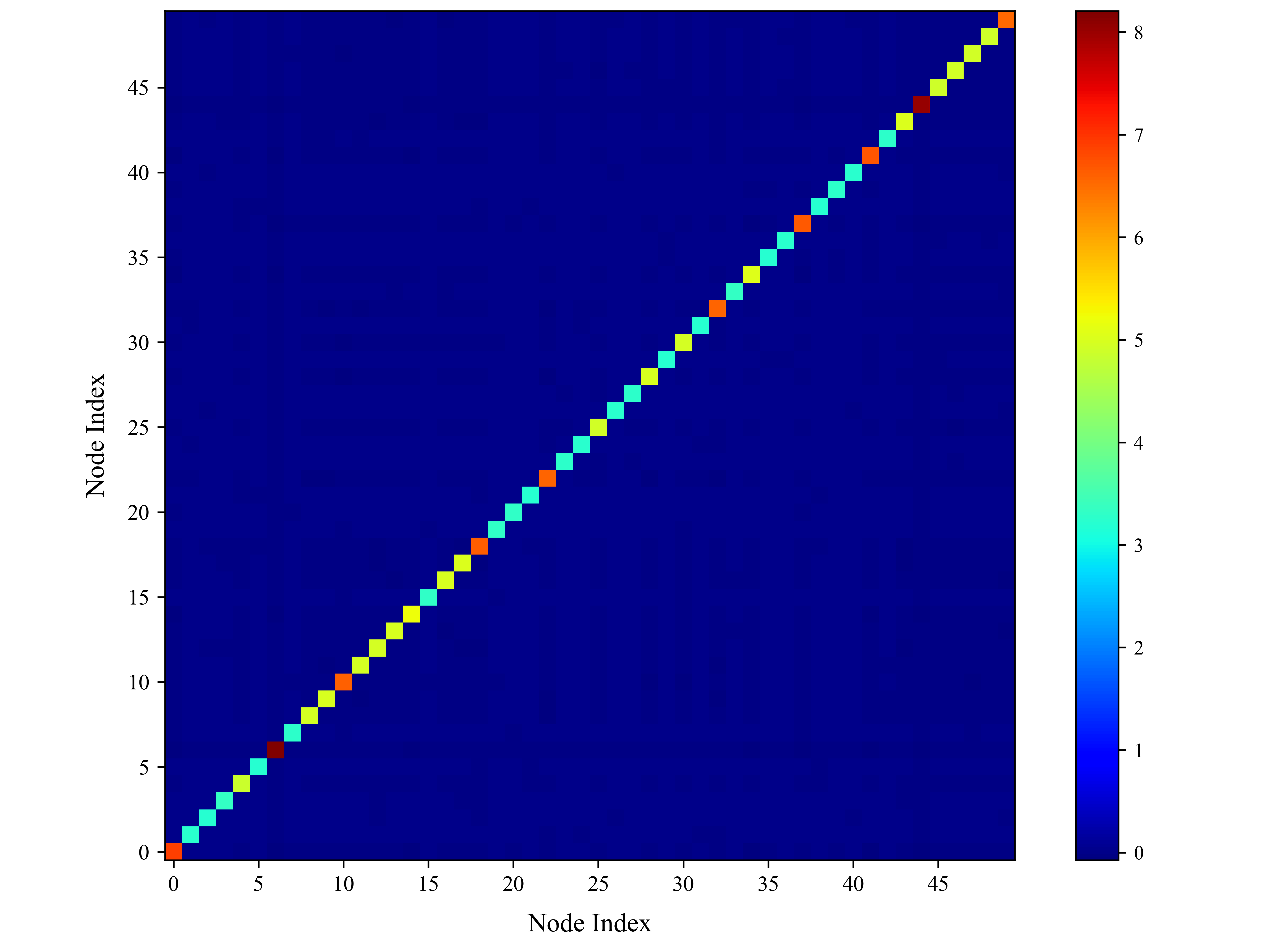}%
		\label{fig_fifth}}
	\hfil
	\subfloat[]{\includegraphics[width=1.7in]{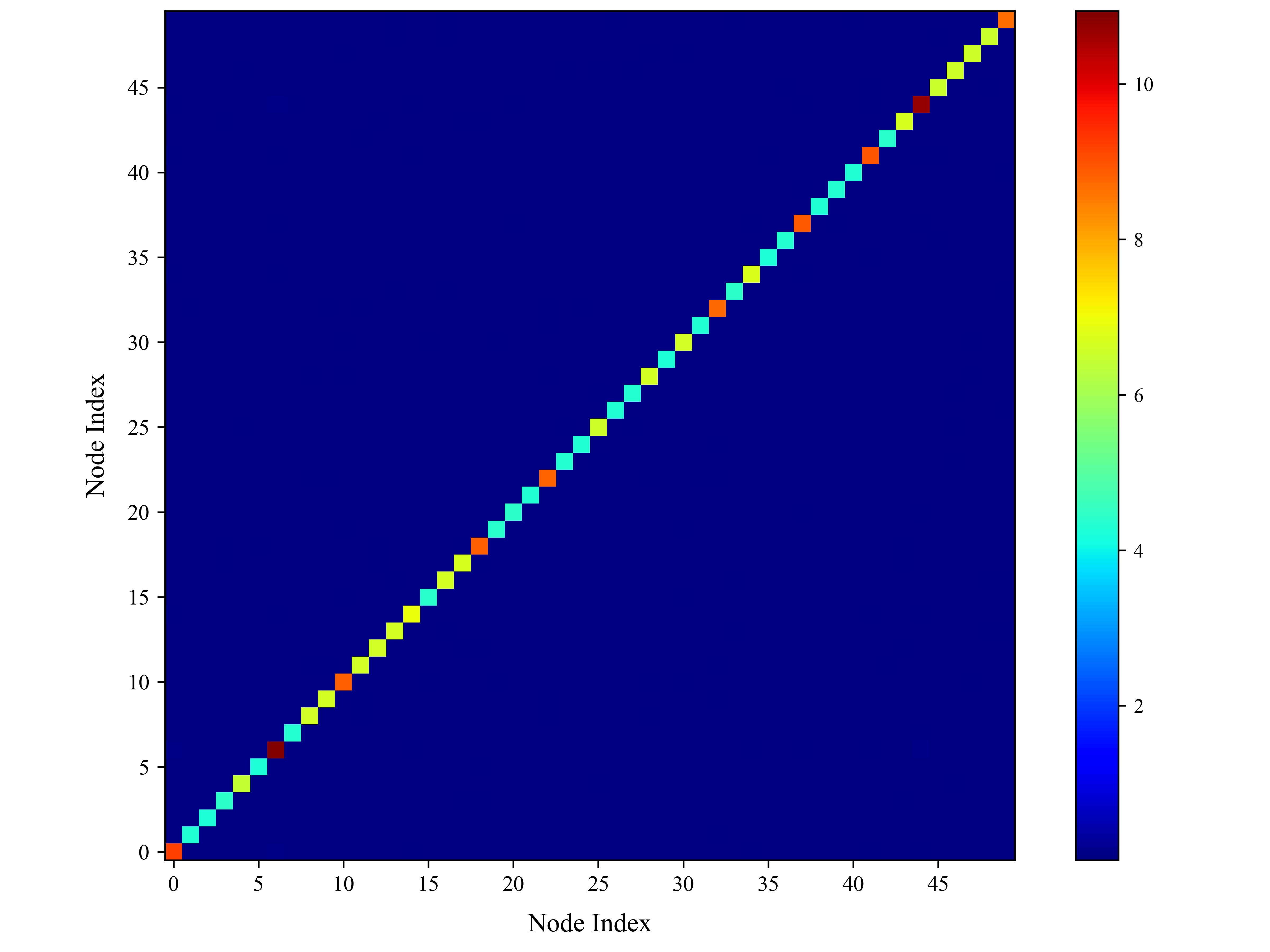}%
		\label{fig_sixth}}
	\caption{Matrix heatmaps for different $m$ and $n$ at $t=1$: (a) $m=0.0$, $n=0.0$ ($\bar{\mathbf{A}}(1)$); (b) $m=0.5$, $n=1.0$ ($\bar{\mathbf{L}}(1)$); (c) $m=1.0$, $n=1.0$ ($\bar{\mathbf{D}}(1)$); (d) $m=0.3$, $n=0.7$; (e) $m=0.6$, $n=0.6$; (f) $m=0.8$, $n=0.2$.}
	\label{fig 10}
\end{figure}

\begin{figure}[!t]
	\centering
	\subfloat[]{\includegraphics[width=1.7in]{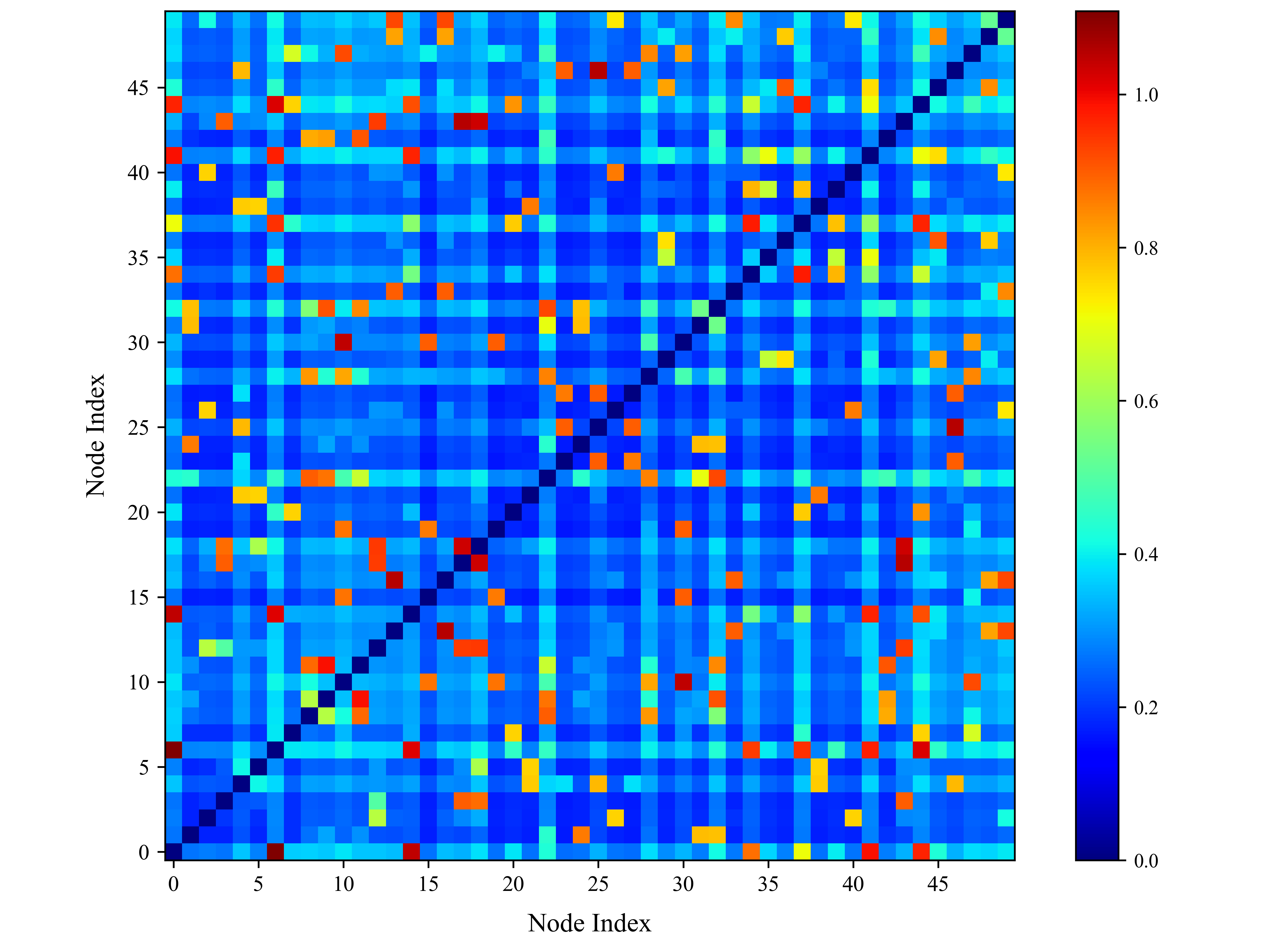}%
		\label{first}}
	\hfil
	\subfloat[]{\includegraphics[width=1.7in]{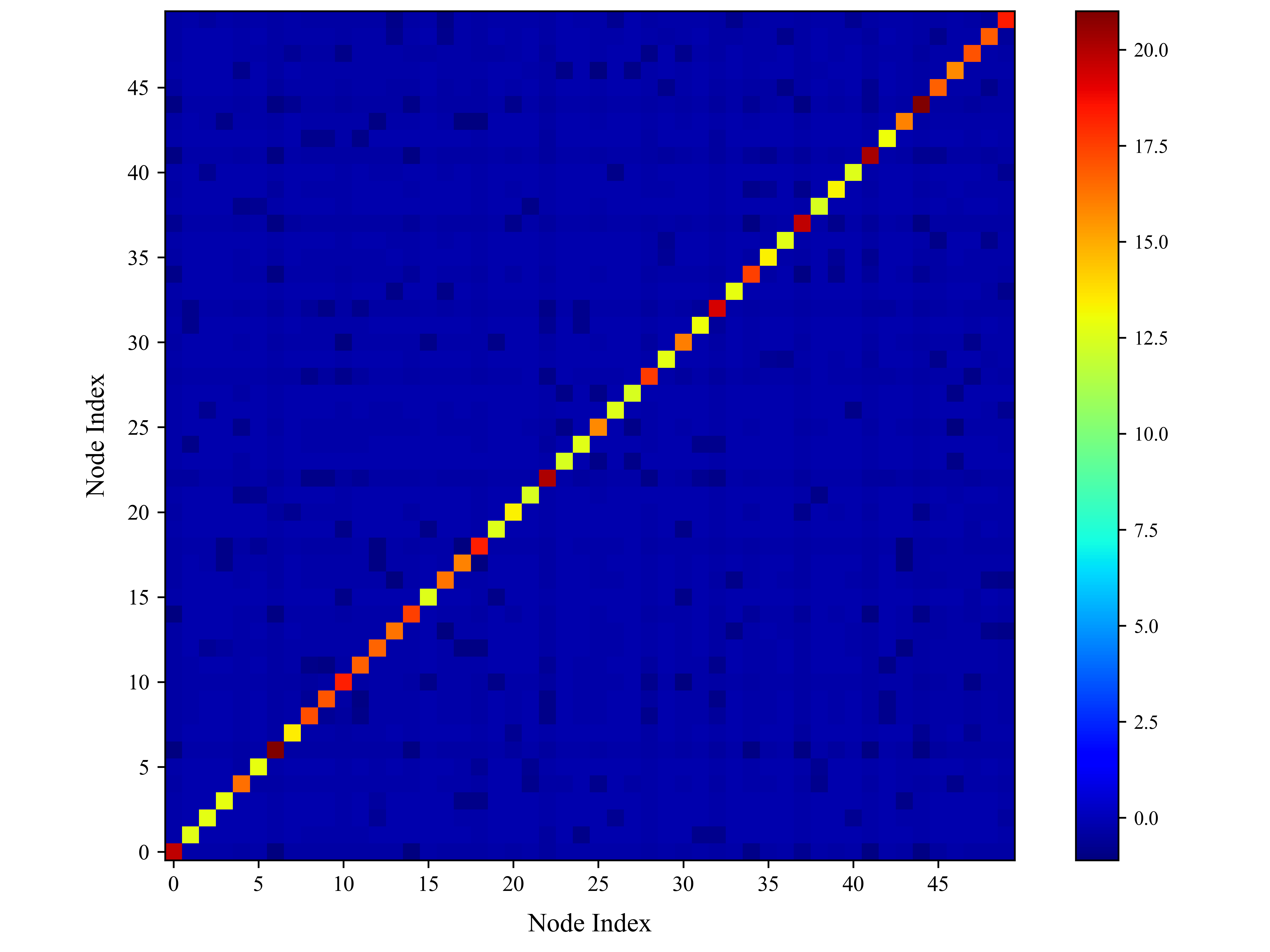}%
		\label{second}}
	\hfil
	\subfloat[]{\includegraphics[width=1.7in]{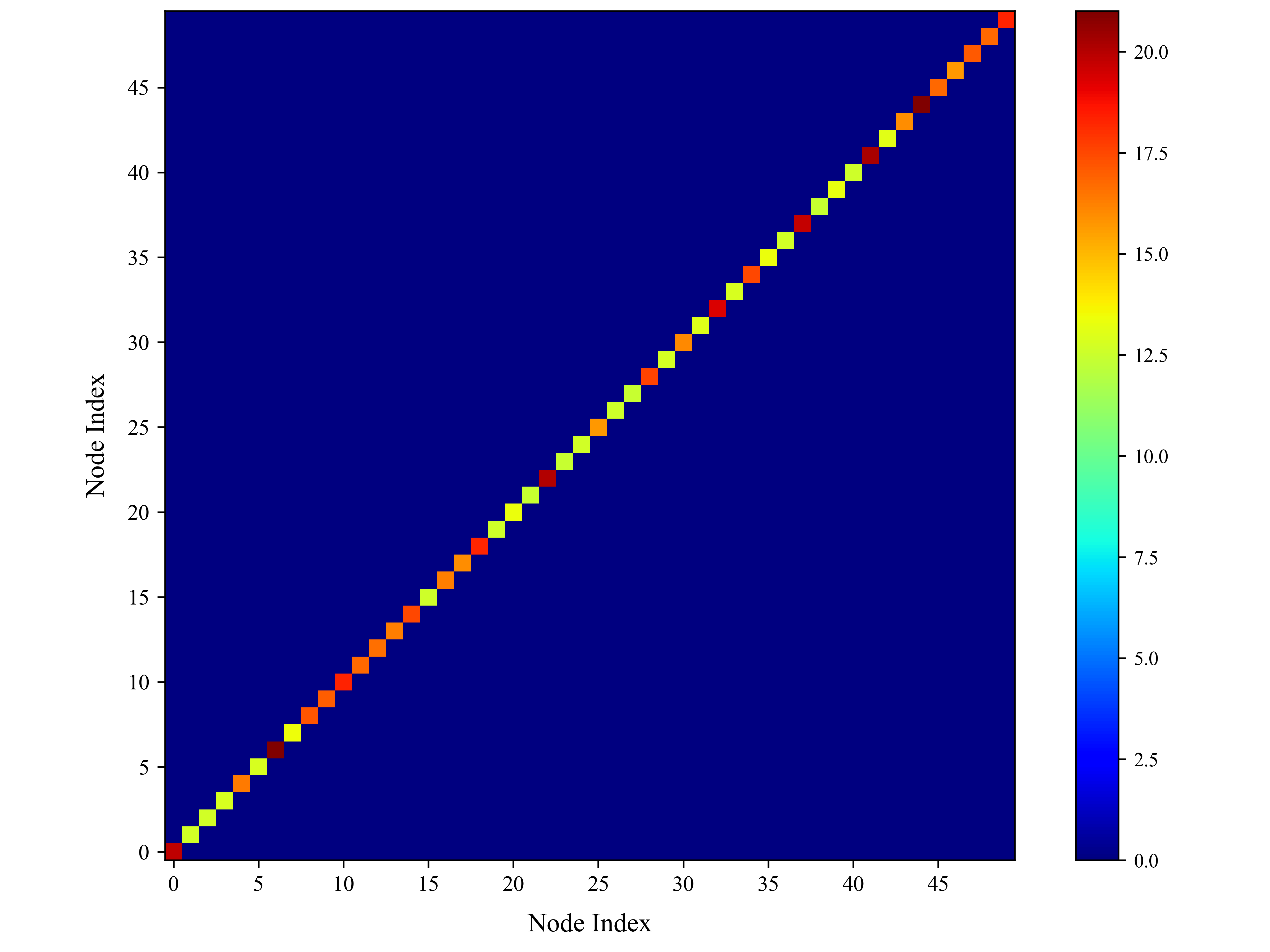}%
		\label{third}}
	\hfil
	\subfloat[]{\includegraphics[width=1.7in]{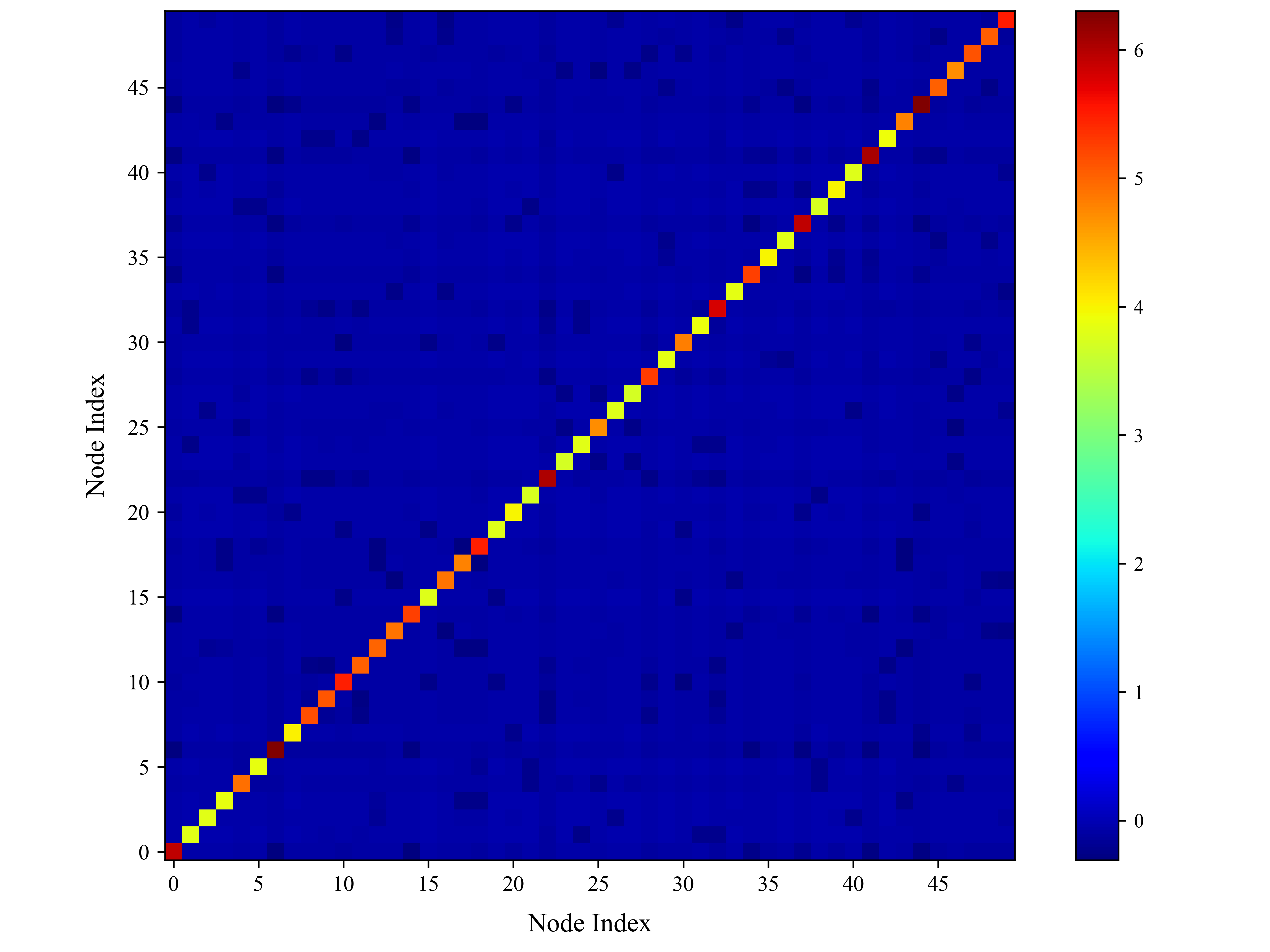}%
		\label{fourth}}
	\hfil
	\subfloat[]{\includegraphics[width=1.7in]{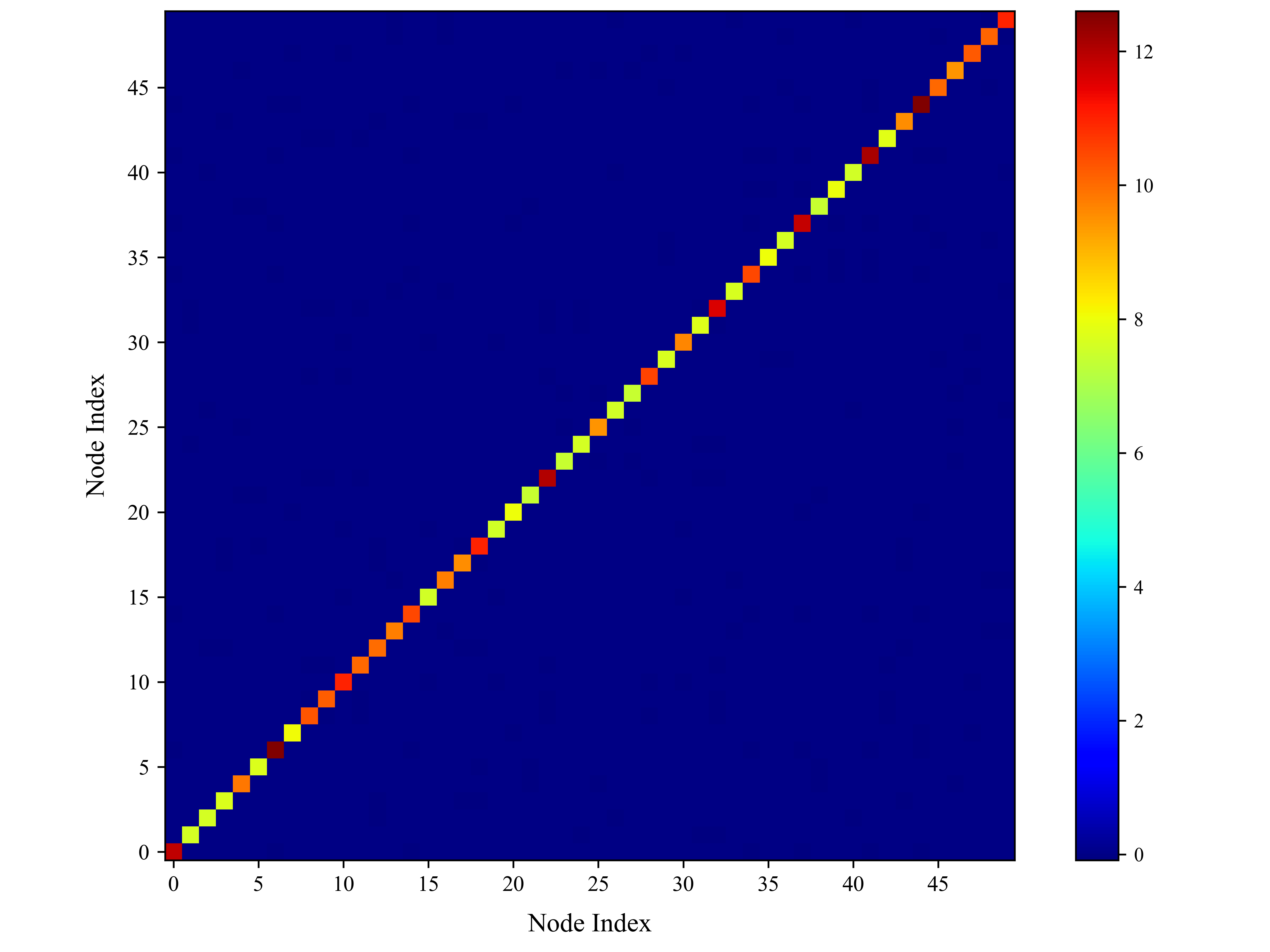}%
		\label{fifth}}
	\hfil
	\subfloat[]{\includegraphics[width=1.7in]{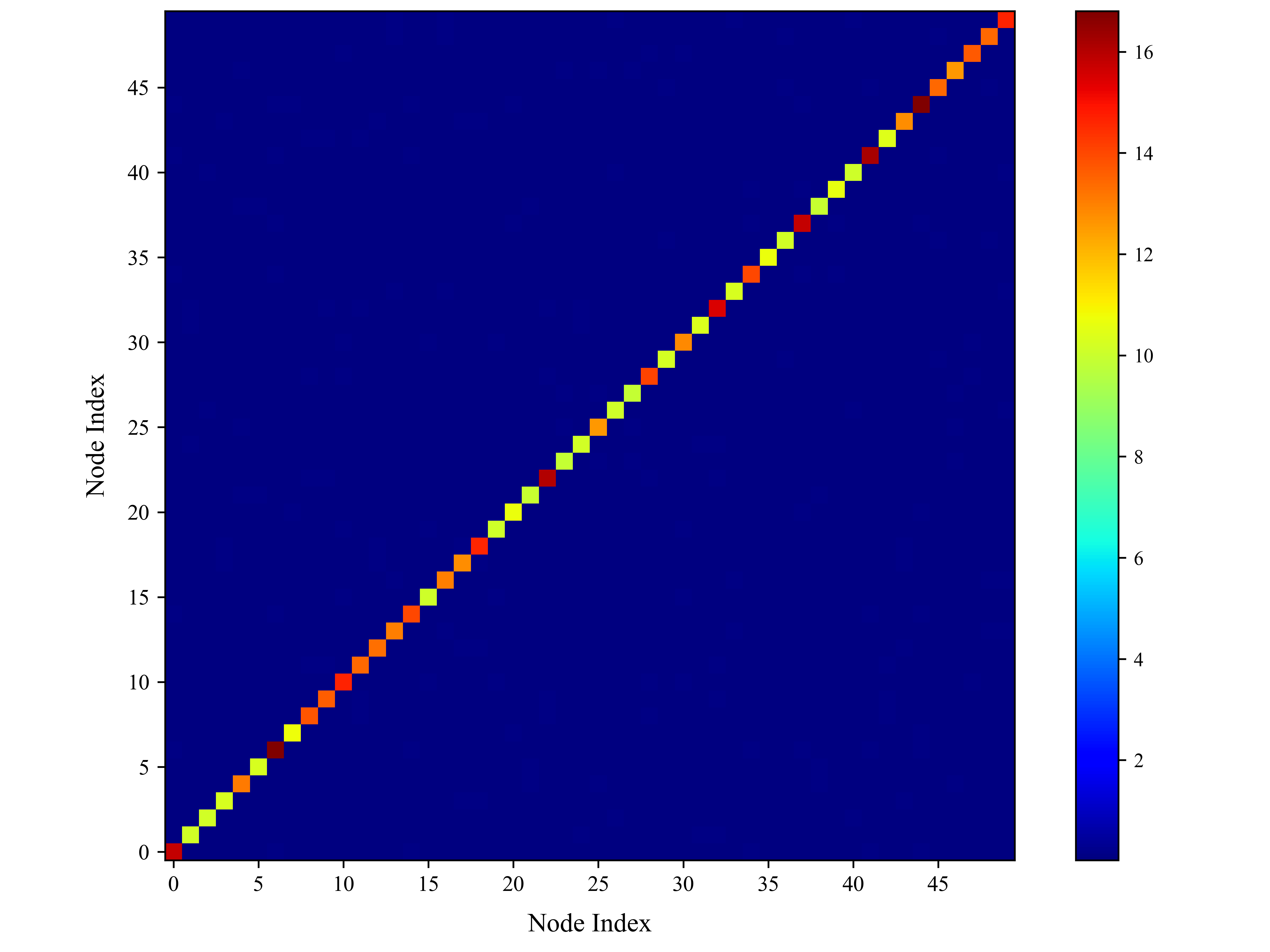}%
		\label{sixth}}
	\caption{Matrix heatmaps for different $m$ and $n$ at $t=2$: (a) $m=0.0$, $n=0.0$ ($\bar{\mathbf{A}}(2)$); (b) $m=0.5$, $n=1.0$ ($\bar{\mathbf{L}}(2)$); (c) $m=1.0$, $n=1.0$ ($\bar{\mathbf{D}}(2)$); (d) $m=0.3$, $n=0.7$; (e) $m=0.6$, $n=0.6$; (f) $m=0.8$, $n=0.2$.}
	\label{fig 11}
\end{figure}

\begin{figure}[!t]
	\centering
	\subfloat[]{\includegraphics[width=1.7in]{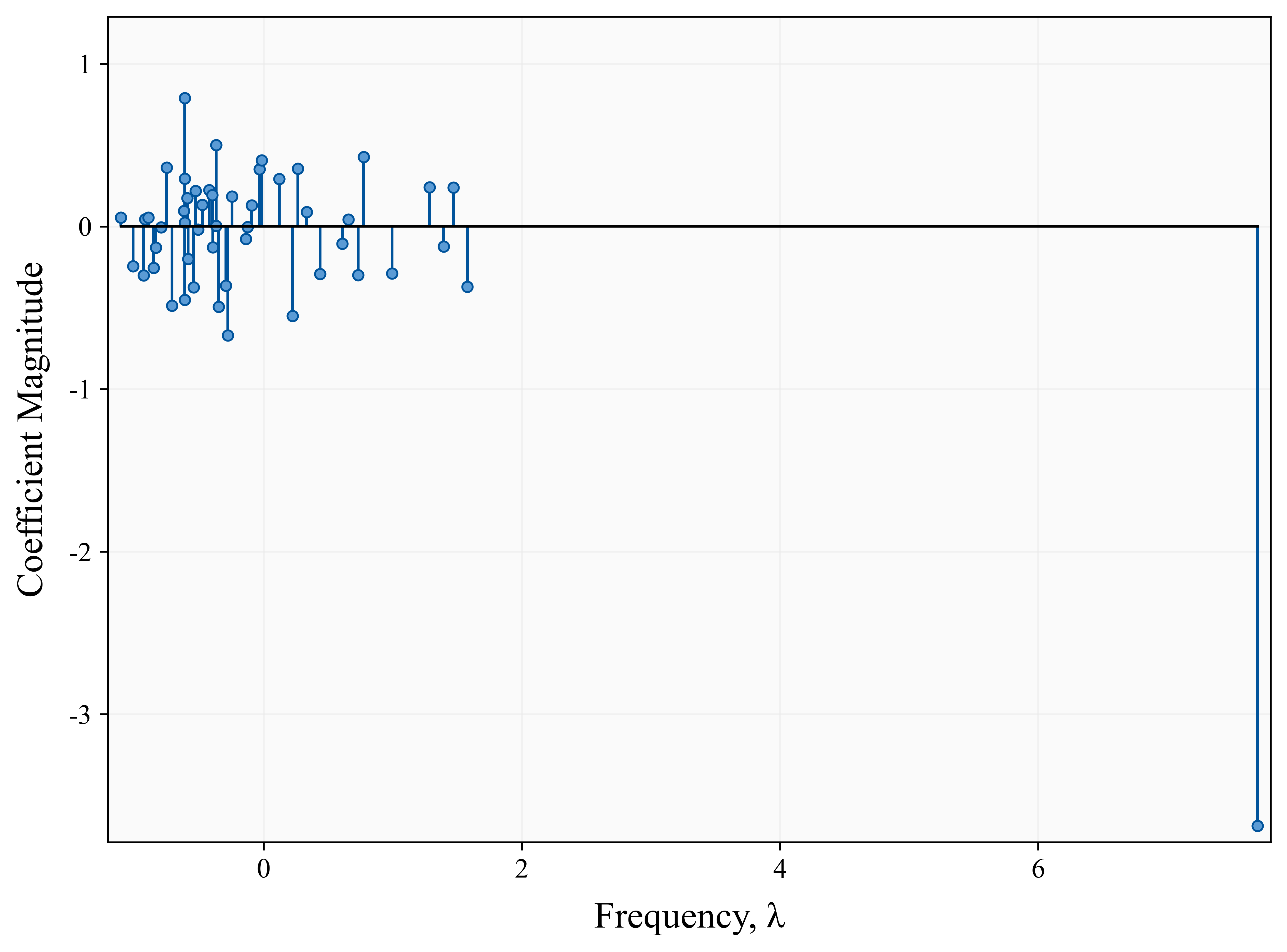}%
		\label{fig_first_case}}
	\hfil
	\subfloat[]{\includegraphics[width=1.7in]{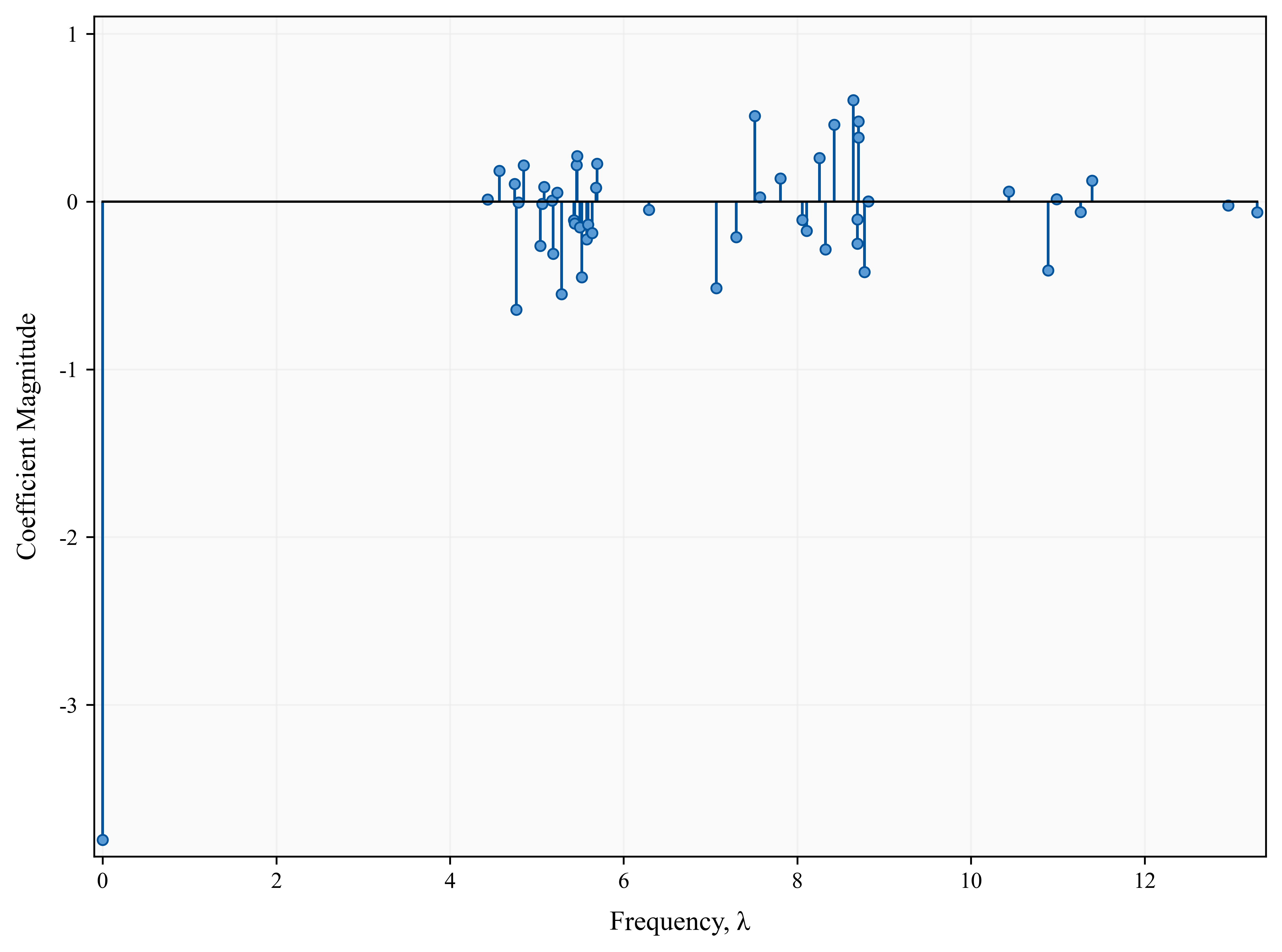}%
		\label{fig_second_case}}
	\hfil
	\subfloat[]{\includegraphics[width=1.7in]{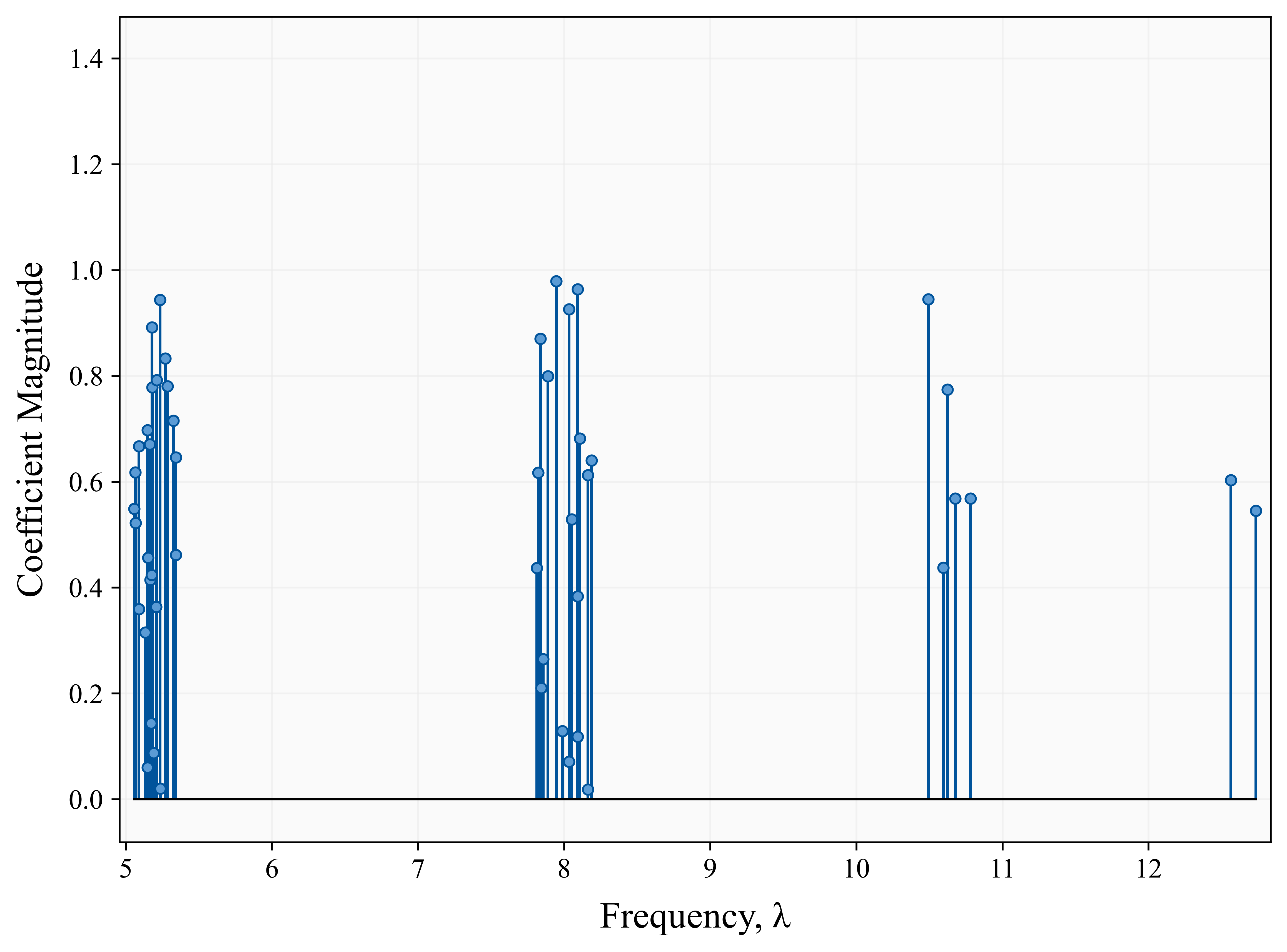}%
		\label{fig_third_case}}
	\hfil
	\subfloat[]{\includegraphics[width=1.7in]{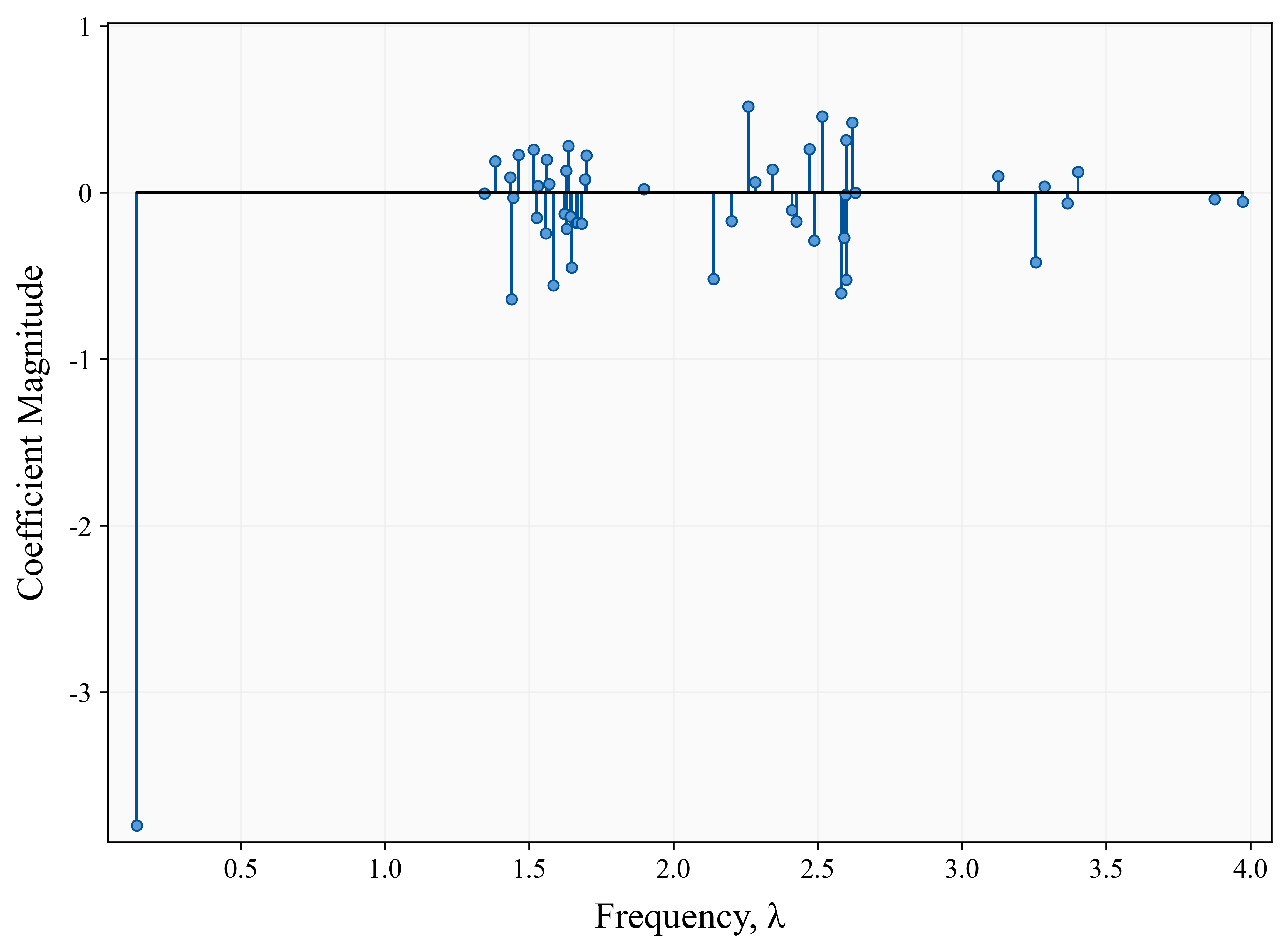}%
		\label{fig_fourth_case}}
	\hfil
	\subfloat[]{\includegraphics[width=1.7in]{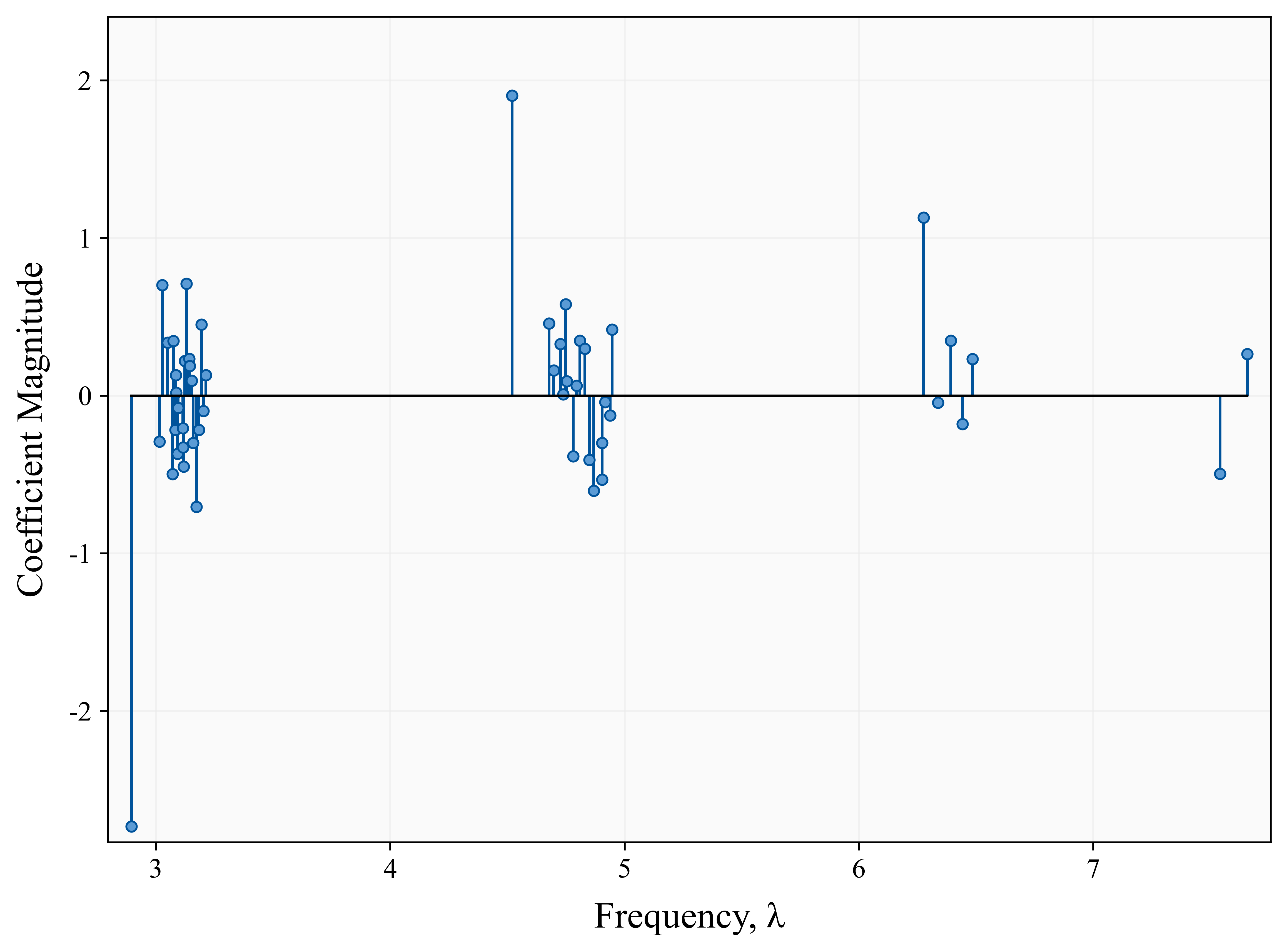}%
		\label{fig_fifth_case}}
	\hfil
	\subfloat[]{\includegraphics[width=1.7in]{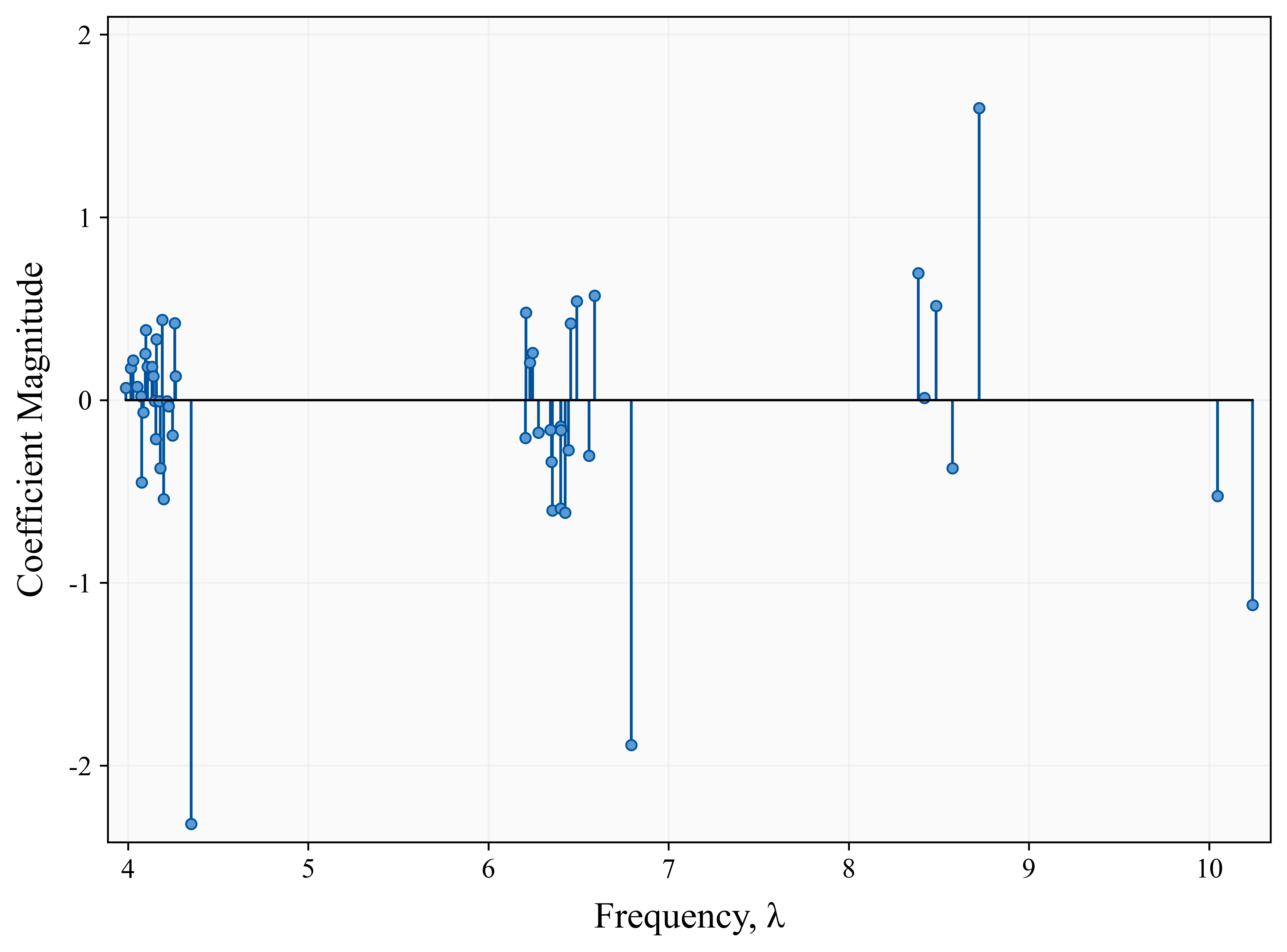}%
		\label{fig_sixth_case}}
	\caption{UEM-GFT of the graph signal $\mathbf{x}$ for different $m$ and $n$ at $t=1$: (a) $m=0.0$, $n=0.0$ ($\bar{\mathbf{A}}(1)$); (b) $m=0.5$, $n=1.0$ ($\bar{\mathbf{L}}(1)$); (c) $m=1.0$, $n=1.0$ ($\bar{\mathbf{D}}(1)$); (d) $m=0.3$, $n=0.7$; (e) $m=0.6$, $n=0.6$; (f) $m=0.8$, $n=0.2$.}
	\label{fig 7}
\end{figure}

\begin{figure}[!t]
	\centering
	\subfloat[]{\includegraphics[width=1.7in]{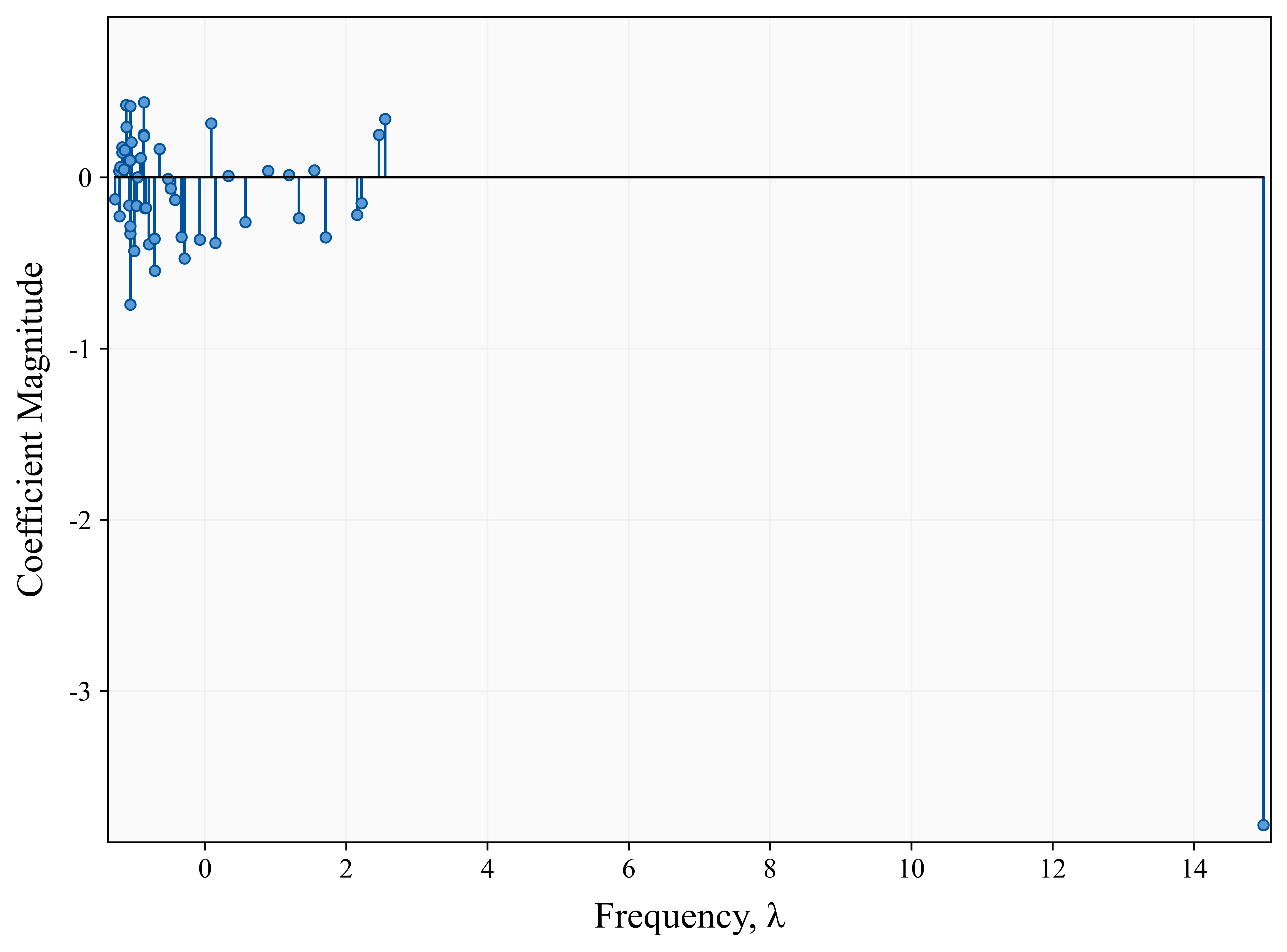}%
		\label{fig_one_case}}
	\hfil
	\subfloat[]{\includegraphics[width=1.7in]{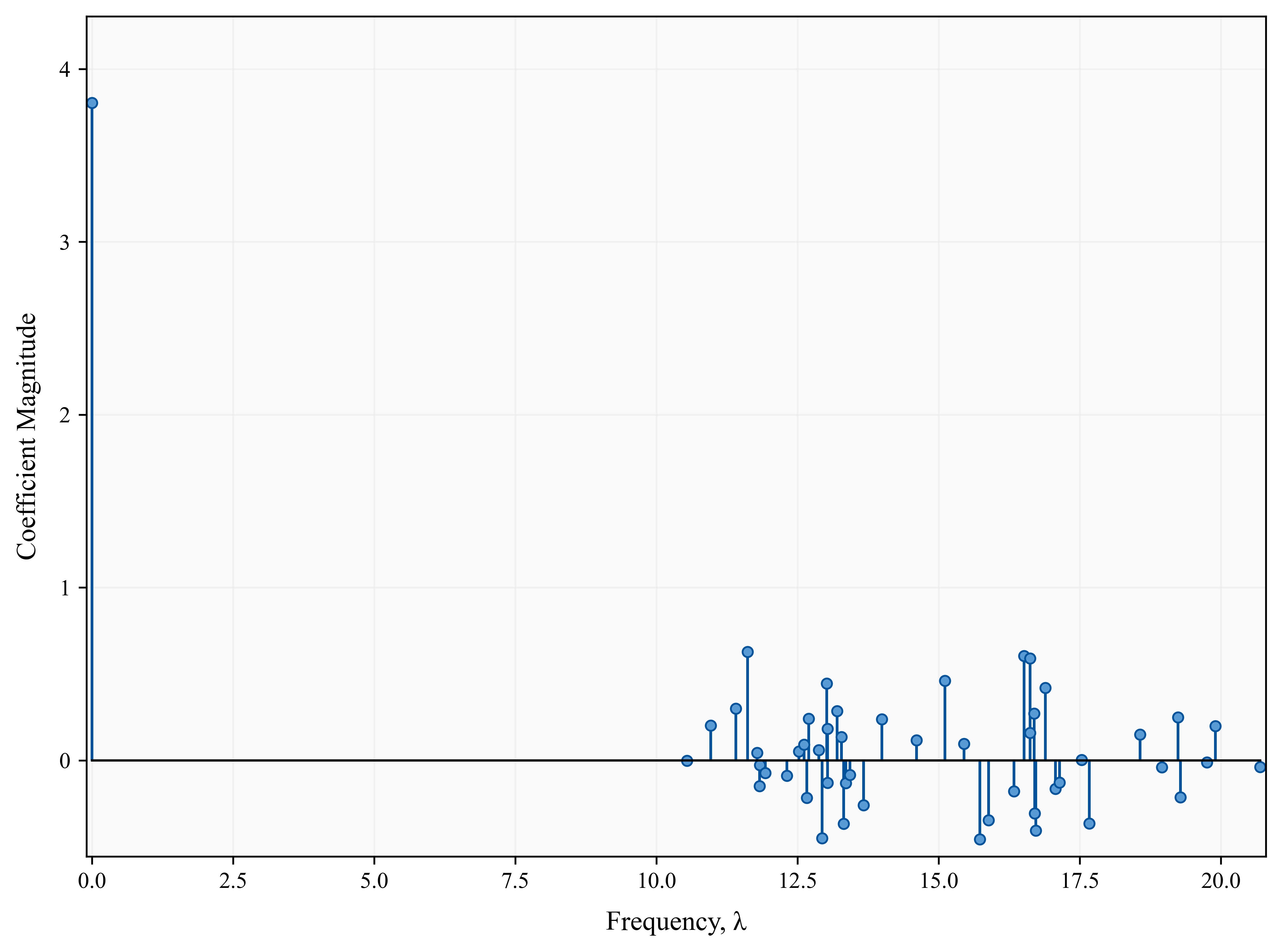}%
		\label{fig_two_case}}
	\hfil
	\subfloat[]{\includegraphics[width=1.7in]{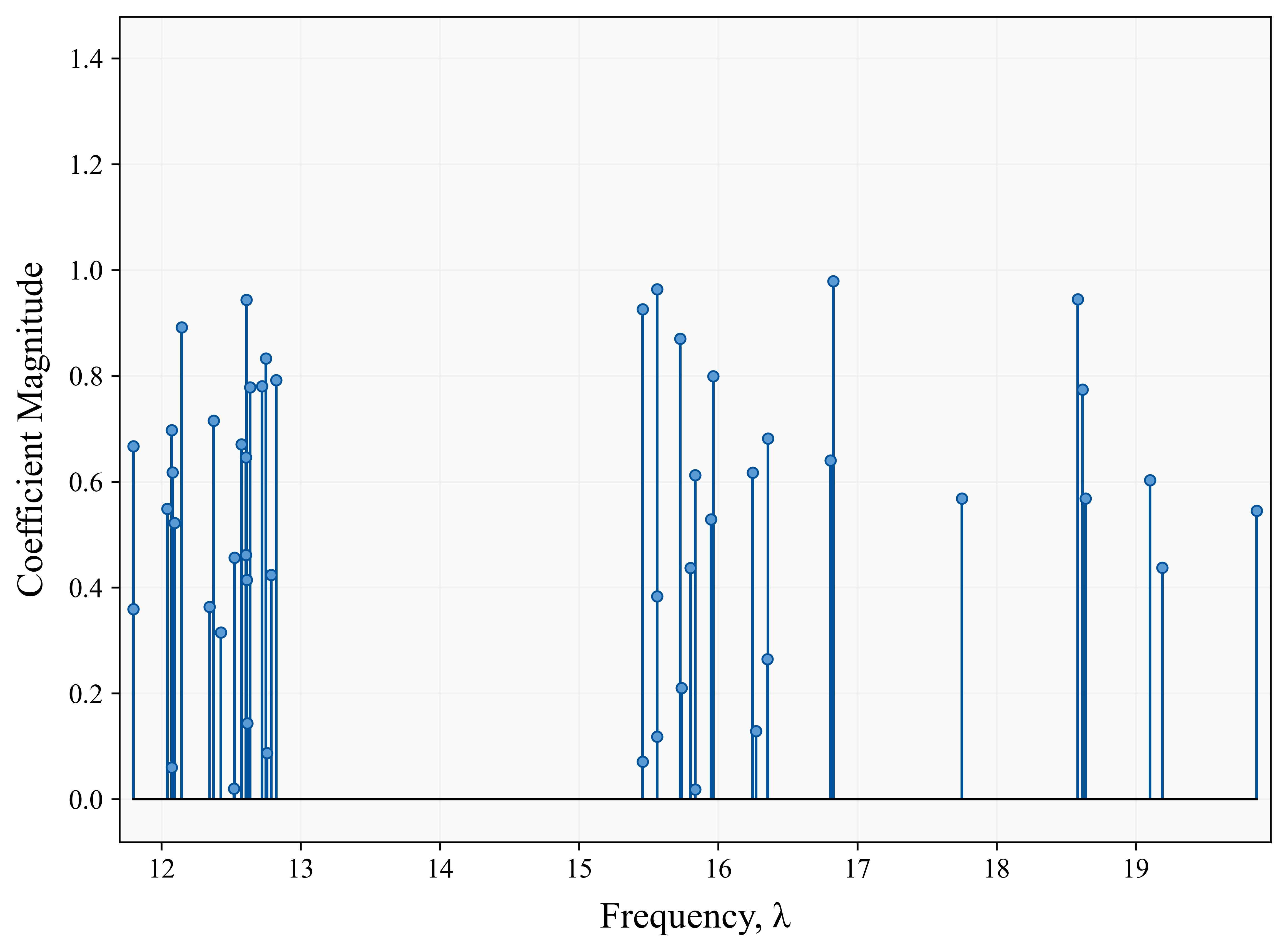}%
		\label{fig_three_case}}
	\hfil
	\subfloat[]{\includegraphics[width=1.7in]{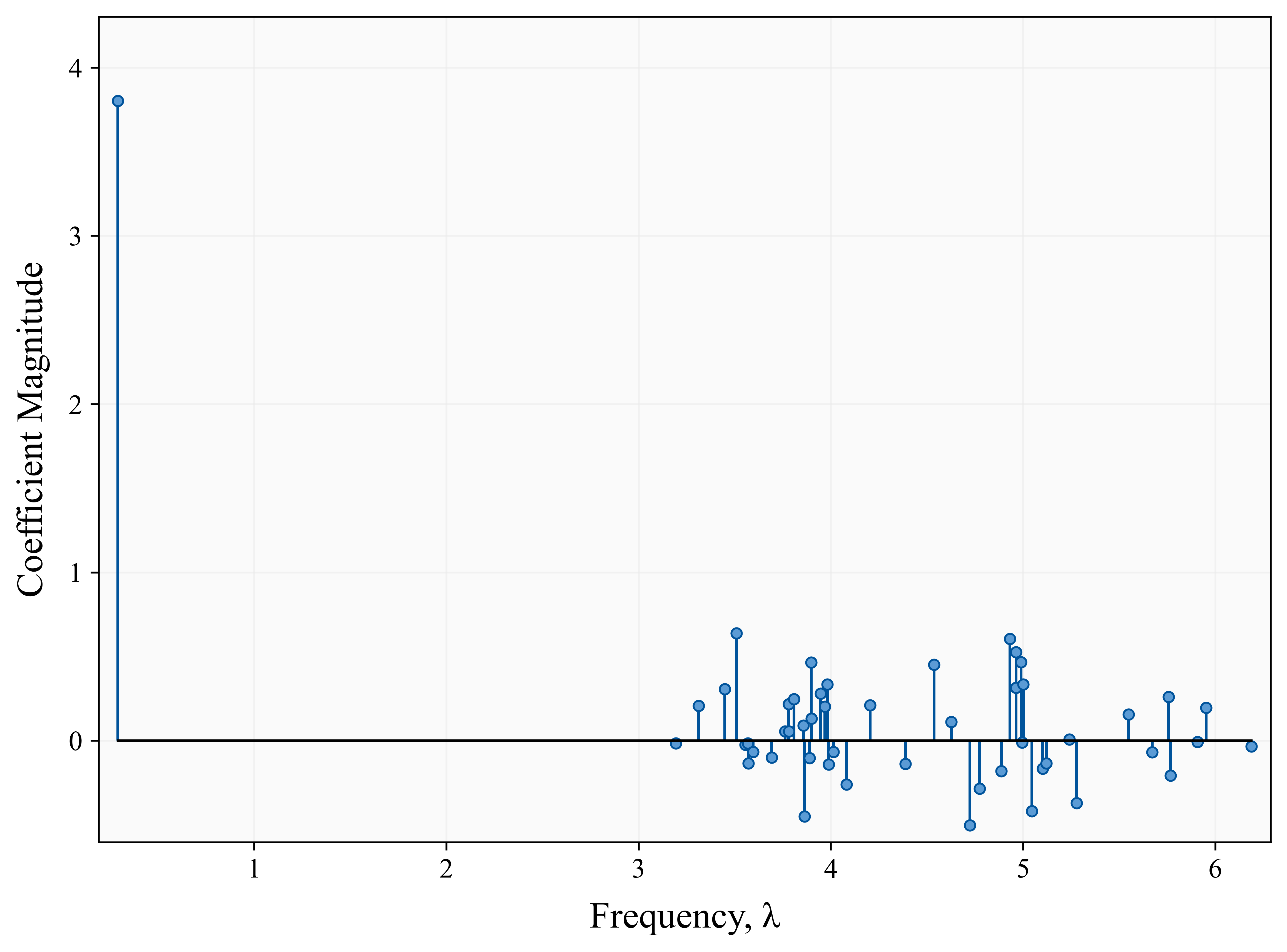}%
		\label{fig_four_case}}
	\hfil
	\subfloat[]{\includegraphics[width=1.7in]{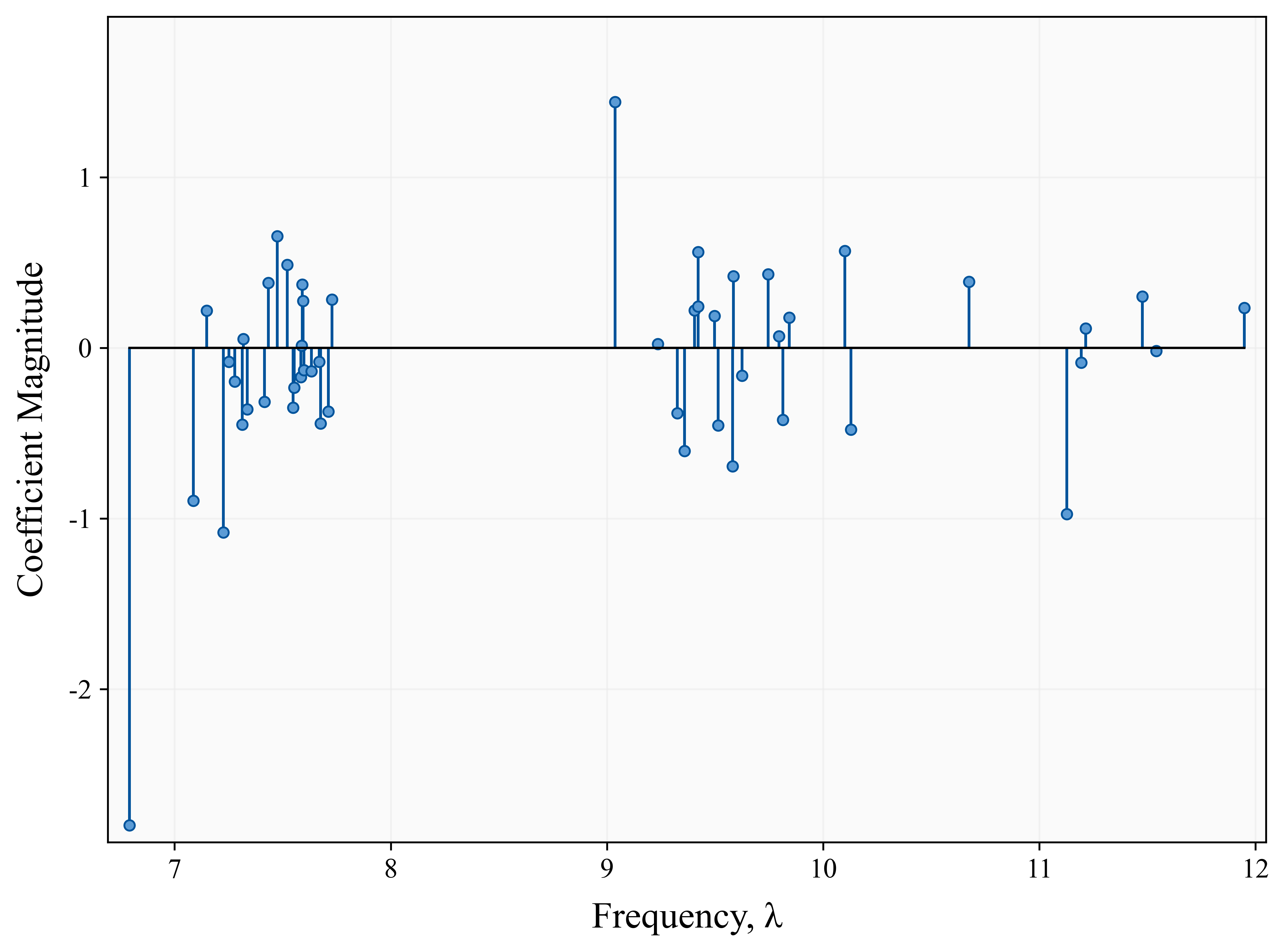}%
		\label{fig_five_case}}
	\hfil
	\subfloat[]{\includegraphics[width=1.7in]{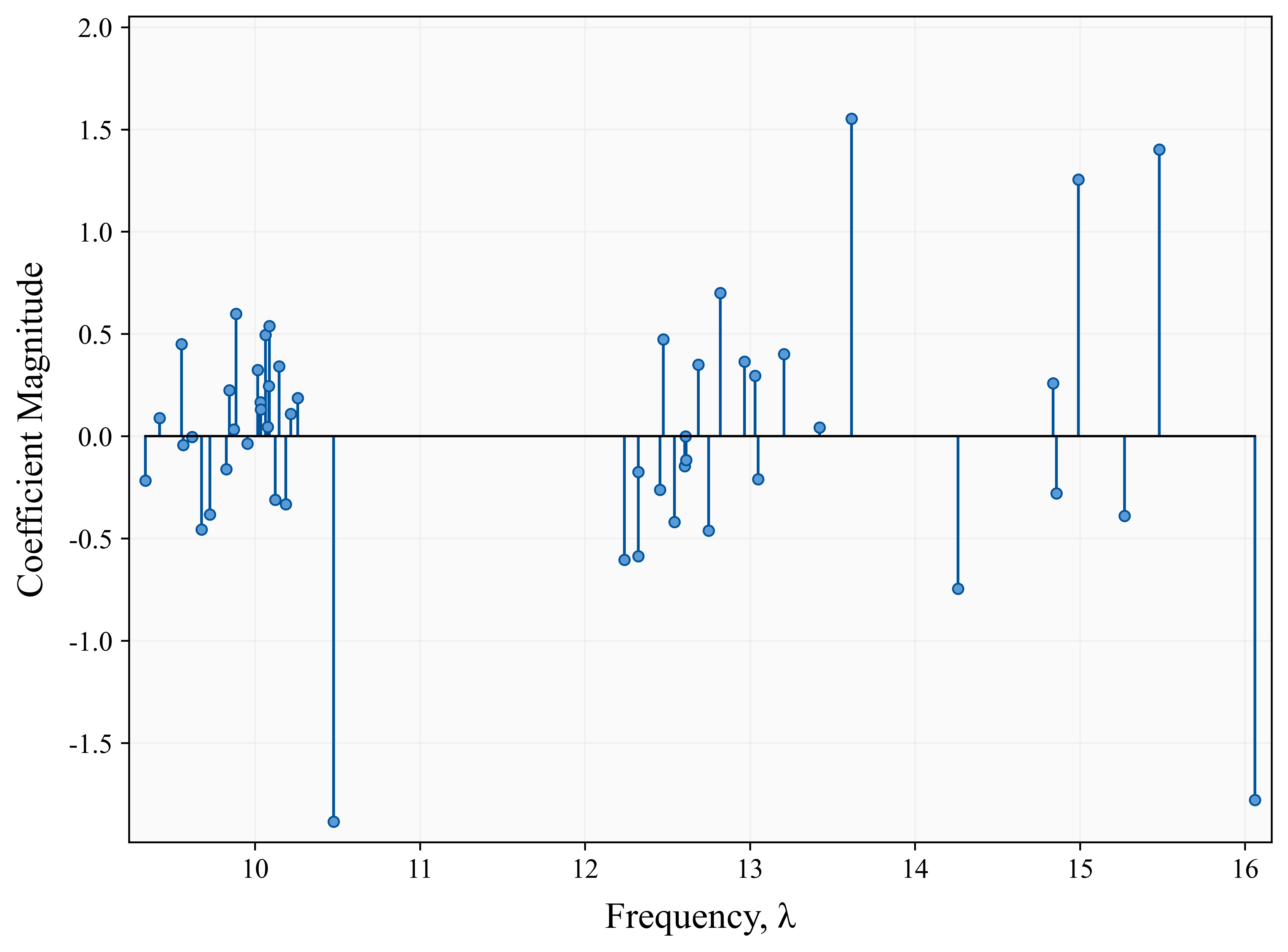}%
		\label{fig_six_case}}
	
	\caption{UEM-GFT of the graph signal $\mathbf{x}$ for different $m$ and $n$ at $t=2$: (a) $m=0.0$, $n=0.0$ ($\bar{\mathbf{A}}(2)$); (b) $m=0.5$, $n=1.0$ ($\bar{\mathbf{L}}(2)$); (c) $m=1.0$, $n=1.0$ ($\bar{\mathbf{D}}(2)$); (d) $m=0.3$, $n=0.7$; (e) $m=0.6$, $n=0.6$; (f) $m=0.8$, $n=0.2$.}
	\label{fig 8}
\end{figure}

Building upon the validated monotonicity of eigenvalues with respect to $m$, we further characterize the parametric evolution of the UEM-GFT by examining a uniformly distributed graph signal $\mathbf{x} \sim \mathcal{U}(0,1)$ on a sensor network. We conduct a sensor graph with $N = 50$ sensors, $k = 3$ and $\rho = 0.3$. Figs. \ref{fig 10} and \ref{fig 11} display heatmaps of UEM for different parameter configurations. As shown in Figs. \ref{fig 7} and \ref{fig 8}, the spectral representations of this signal under varying parameters $m$, $n$ and $t$ are demonstrated. Through parametric adjustment of the UEM, the proposed framework captures a broader spectral space and achieves enhanced flexibility in spectral characteristics.


\section{Application}
In this section, we perform comprehensive experiments on both synthetic and real-world datasets. The goal is to compare the performance of anomaly detectors developed from our proposed framework against those based on alternative GSO approaches. The UEM provides enhanced flexibility in tuning graph spectral properties via parameter selection, which has the potential to boost anomaly detection performance. Additionally, we select the “best” matrix from the UEM at diffusion scales $t\in\{1, 2\}$.

\subsection{Anomaly Detection Task}
The application of anomaly detection \cite{ref42}–\cite{ref44} is driven by the growing connectedness of modern networked systems, which necessitates robust security mechanisms and fault-tolerant reliability guarantees. Leveraging graph-spectral information produced through UEM-GFT and alternative GSO-based approaches, we develop classification models. For comparative evaluation, the implemented GSO-based approaches include the conventional GFT using the Laplacian matrix, sGFT with diffusion scale $t=1$ (DF1) and $t=2$ (DF2), GFT based on shortest-path-based GSOs with maximum path lengths of 2 hops (SP2) and 3 hops (SP3), and GFT using the Markov matrix (MRK) \cite{ref22}. And our current work is to consider combining the extended-adjacency matrix and the unified graph representation matrix to form UEM, so as to find more suitable GSOs for different graph structures and data. 

The method for developing the anomaly detector closely follows prior works \cite{ref12}, \cite{ref39}, \cite{ref45}, \cite{ref46}, as show in Algorithm \ref{alg:UEM-GFT}. The implementation is carried out in Python using scikit-learn's GridSearchCV class. Specifically, the detection framework is established on the fundamental assumption that healthy signals exhibit spectral smoothness characteristics. A high-pass filtering operation with cutoff frequency $\lambda_{\text{cut}}$ is implemented to extract high-frequency components, upon which the classification procedure is performed. Let denote the training dataset $\mathcal{X}_{\text{train}} = \{\mathcal{X}_H, \mathcal{X}_A\}$, where $\mathcal{X}_H$ represents the subset of healthy graph signals (anomaly-free instances) and $\mathcal{X}_A$ comprises anomalous graph signals. The discrimination between healthy and anomalous signals is achieved through threshold-based detection, where any signal containing spectral coefficients exceeding the predetermined threshold $\tau$ is identified as anomalous. The detection threshold $\tau$ is obtained through this procedure:
\begin{enumerate}
	\item{compute coefficients in the graph-frequency domain for every healthy signal within $\mathcal{X}_H$;}
	\item{perform filtering that preserves only components associated with graph frequencies above $\lambda{\text{cut}}$;}
	\item {for each signal, a partial $\tau_p$ corresponds to the maximum absolute coefficient after filtering;}
	\item{generate the final detection threshold using all computed $\tau_p$ values:}
\end{enumerate}
\begin{equation}
	\tau = \mu_{\tau_p} + \beta\sigma_{\tau_p},
\end{equation}
where $\mu_{\tau_p}$ and $\sigma_{\tau_p}$ denote the estimated mean and standard deviation, respectively, of the partial thresholds $\tau_p$. The non-negative parameter $\beta$ scales the confidence level relative to this standard deviation.

For GFT implementations (using either Laplacian matrix or Markov matrix), only two hyperparameters require training: $\lambda_{\text{cut}}$ and $\beta$. In contrast, methods with scale dependence require extra parameters: length of the shortest-path, plus diffusion scale $t$ and normalization parameter $\rho$ for sGFT. We independently examine two distinct diffusion scales $t\in\{1, 2\}$ and maximum path lengths of 2 and 3. Furthermore, for UEM-GFT, we maintain the same diffusion scales $t\in\{1, 2\}$, and the other two parameters $m$ and $n$ of the plan $\bar{\mathbf{P}}_{m,n}(t)$ are taken between 0 and 1 by step 0.1. To ensure a fair comparison, hyperparameters $\lambda_{\text{cut}}$, $\beta$ and $\rho$ are optimized via grid-search with consistent ranges and step sizes across all methods. In essence, a predefined set of values is assigned to each hyperparameter, and all possible combinations are meticulously evaluated based on a specific metric. We adopt the F1 score as metric, which is given by:
\begin{equation}
	\text{F1} = \frac{2 \times \text{TP} }{2 \times \text{TP} + \text{FP} + \text{FN}},
\end{equation}
with TP denoting correctly identified anomaly, FP representing healthy signal incorrectly flagged as anomaly, and FN indicating anomaly mistakenly classified as healthy signal.

Additionally, hyperparameter optimization employs 5-fold cross-validation, partitioning the training set into five equal subsets. For each hyperparameter combination, the model is trained (computing $\tau$) over 4 sets and evaluated over the remaining set. The optimal hyperparameter set is selected based on the best average result all folds. Once the classifiers are trained, their performance is rigorously evaluated by calculating the F1 score on a separate test dataset, denoted as $\mathcal{X}_{\text{test}}$. To ensure the reliability of our results, all experiments present simulation results averaged over 50 distinct randomly generated training and test pairs. The specific configurations of these datasets are tailored to align with the requirements of each simulation.

To construct $k$-NN graphs for both synthetic and real-world datasets, we utilize the pygsp library, varying the number of nodes $N \in \{10,30,50\}$ and neighbor nodes $k \in \{3, 6\}$. For the synthetic datasets, we generate   graphs by distributing $N$ sensors uniformly within the unit square space $[0, 1] \times [0, 1]$. However, for real-world datasets, stations $N$ selected randomly from the available stations and the network structure of the $k$-NN graphs are then determined based on the geographical coordinates (latitudes and longitudes) of these selected stations.

For real-world datasets, it is assumed that all initially data available are healthy. Therefore, anomaly must be artificially injected into the data for the purpose of experimentation. In Experiment 2 involving synthetic datasets, the method for constructing anomaly is identical to that employed for real-world datasets. Specially, anomaly is introduced via additive Gaussian noise over healthy data. The noise's average follows a discrete uniform distribution, taking non-zero integer values within $[-b_{\text{max}}, +b_{\text{max}}]$. Additionally, the maximum number of anomalous sensors is set beforehand and explicitly stated for each simulation.
\renewcommand{\algorithmicrequire}{\textbf{Input:}}
\renewcommand{\algorithmicensure}{\textbf{Output:}}
\begin{algorithm}[t]
	\caption{Anomaly Detection Based on UEM-GFT}\label{alg:UEM-GFT}
	\begin{algorithmic}[1]
		\REQUIRE 
		~\\ Graph signal, An $ N \times N$ matrix $\mathbf{A}$, Parameters $m, n, t$,
		~\\ Hyperparameters $\rho$, $\lambda_{\text{cut}}$ and $\beta$.
		\ENSURE 
		~\\ Mean F1 score over 50 runs.
		\STATE Construct extended-adjacency matrix $\bar{\mathbf{A}}(t)$.
		\STATE Construct UEM:
		$\bar{\mathbf{P}}_{m,n}(t) := m\bar{\mathbf{D}}(t) + (2n - 1)(m - 1)\bar{\mathbf{A}}(t)$.
		\STATE Generate healthy signals and anomalous signals, and split into train datatest $\mathcal{X}_{\text{train}}$ and test datatest $\mathcal{X}_{\text{test}}$, where $\mathcal{X}_{\text{train}} = \{\mathcal{X}_H, \mathcal{X}_A\}$, $\mathcal{X}_H$ denotes healthy signals and $\mathcal{X}_A$ denotes anomalous signals.
		\STATE Compute the detection threshold $\tau = \mu_{\tau_p} + \beta\sigma_{\tau_p}$ using $\mathcal{X}_H$.
		\STATE Calculate F1 score on $\mathcal{X}_{\text{test}}$.
		\STATE Repeat steps 3-5 for 50 runs and compute mean F1 score.
	\end{algorithmic}
\end{algorithm}

\subsection{Simulations Over Synthetic Networks}

\begin{figure}[!t]
	\centering
	\subfloat{\includegraphics[width=1.7in]{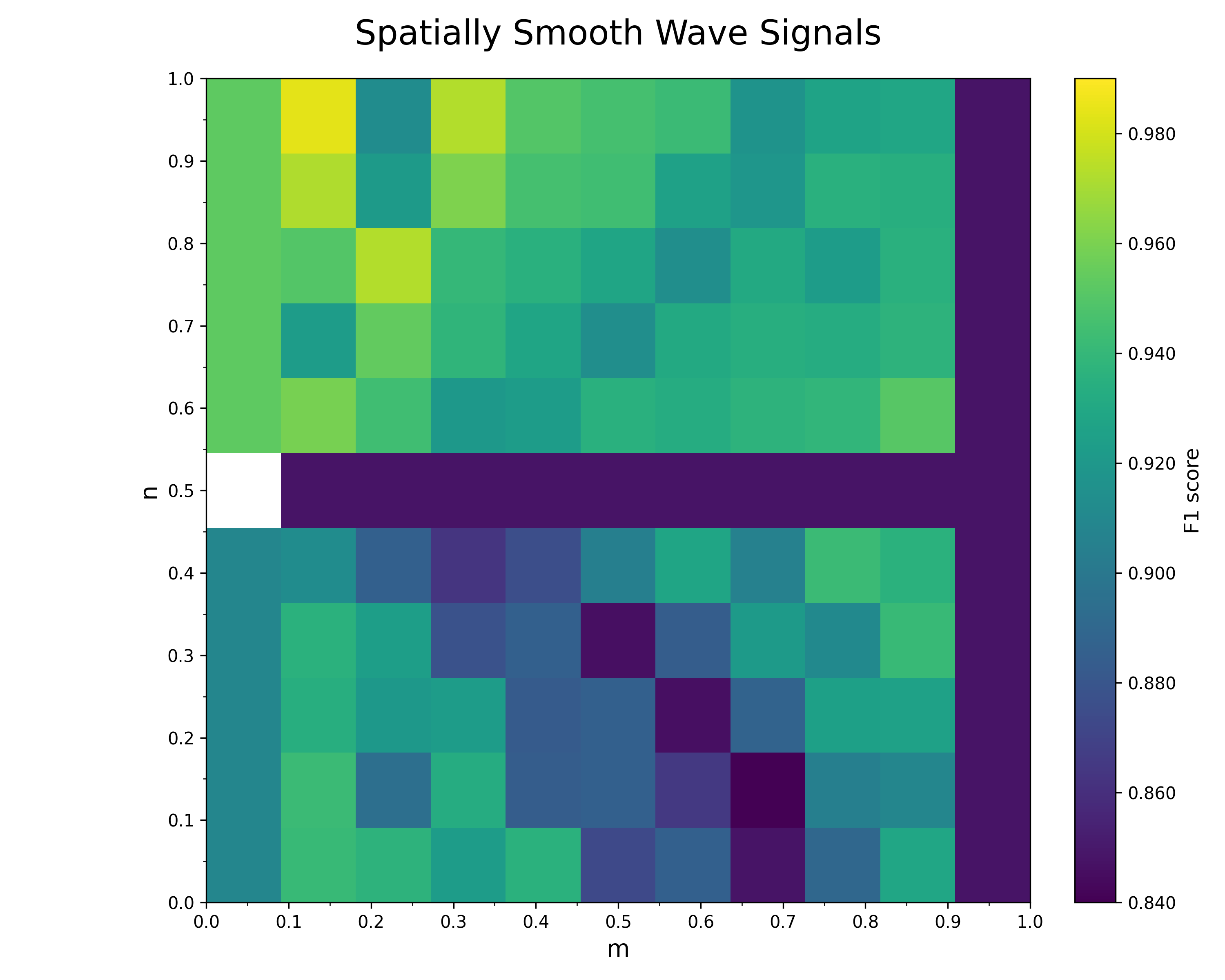}}
	\subfloat{\includegraphics[width=1.7in]{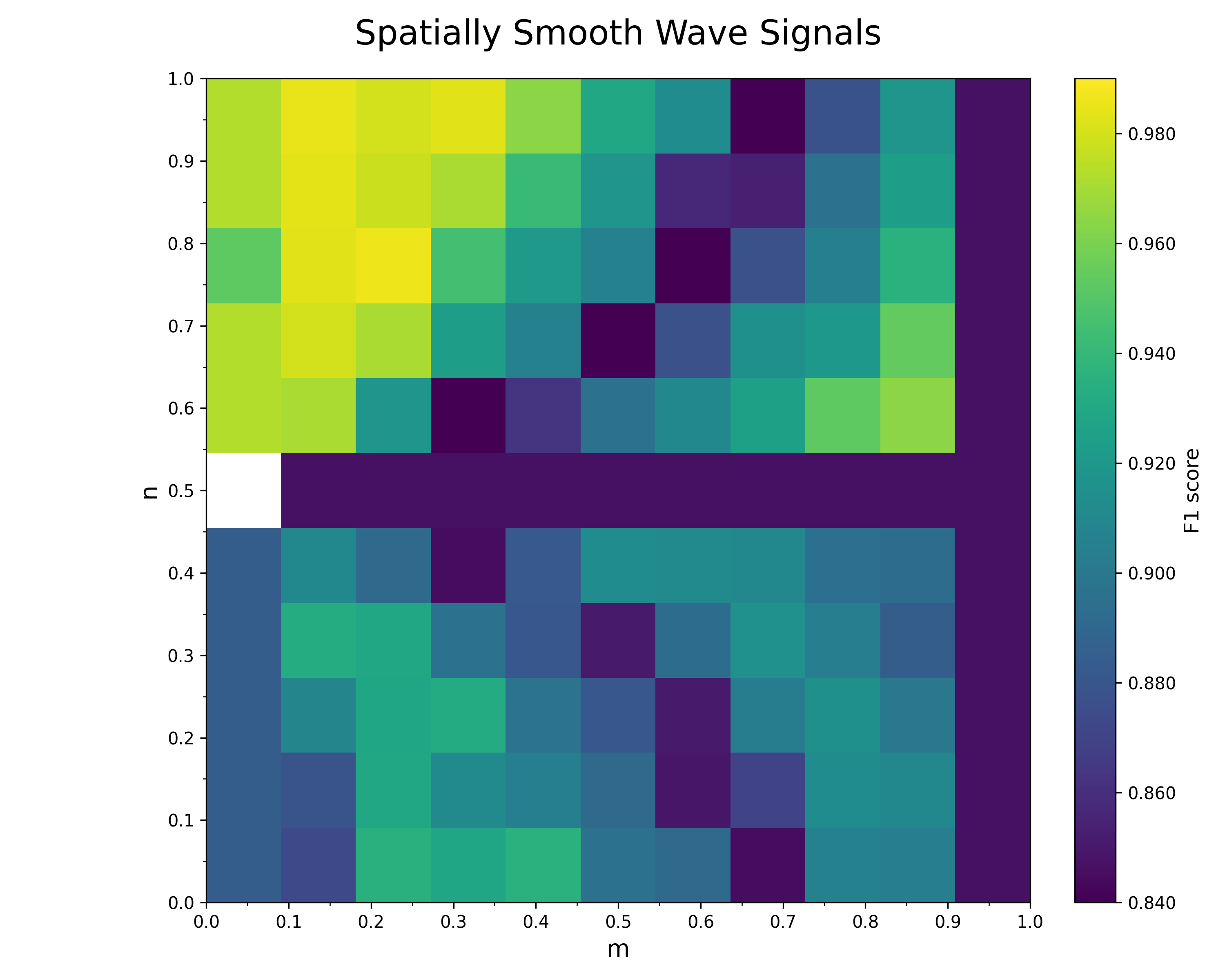}}
	\caption{Figures (Left: $t=1$/Right: $t=2$) depict F1 scores for Spatially Smooth Wave Signals and for different values of $m$ and $n$ defining $\bar{\mathbf{P}}_{m,n}(t)$ with $N=30$, $k=3$ and $t \in \{1, 2\}$.}
	\label{fig 2}
\end{figure}

\textit{1) Experiment 1:} Spatially Smooth Wave Signals - spatially-spread anomaly: The sensors capture a spatially smooth wave signal given by $s(d_x, d_y) = \cos(2\pi d_x + \theta_x) + \cos(2\pi \cdot 2 d_y + \theta_y)$, where $d_x, d_y \in [0, 1]$ represent horizontal and vertical spatial coordinates, and $\theta_x$, $\theta_y$ denote dynamic phase offsets for the signal in the $x$-direction and $y$-direction, respectively. The initial value of the phase values is $0$ and they are updated at $[-0.5, 0.5]$ with uniform sampling by a step factor of $0.1$ and $0.05$, respectively, and each update is independent of each other. We address the detection of an additive high-frequency interference signal given by $s_i(d_x, d_y) = 0.1 \left( \cos(2\pi \cdot 5d_x + \theta_x) + \cos(2\pi \cdot 6d_y + \theta_y) \right)$. Datasets for training $\mathcal{X}_{\text{train}}$ and testing $\mathcal{X}_{\text{test}}$ each contain $150$ healthy samples and $150$ anomalous instances. As depicted in Fig. \ref{fig 2}, “maps” that represent F1 scores are different from one diffusion scale to another. The F1 score results obtained across the 50 independent runs are summarized in Table \ref{table 1}. The experimental results show that the detector based on the extended-Laplacian matrix outperforms the detectors with other GSOs. In addition, the detection performance can be further improved by introducing UEM.
\begin{table*}
	\centering
	\caption{F1 scores of different GSOs under different nodes $N$ and  neighboring nodes $k$ for Spatially Smooth Wave Signals.}
	\begin{tabular}{@{}l@{\hspace{5em}}c@{\hspace{5em}}c@{\hspace{5em}}c@{\hspace{5em}}c@{\hspace{5em}}c@{\hspace{5em}}c@{}}
		\toprule
		& \multicolumn{2}{c}{$N=10$} & \multicolumn{2}{c}{$N=30$} & \multicolumn{2}{c}{$N=50$} \\
		\cmidrule(lr){2-3} \cmidrule(lr){4-5} \cmidrule(lr){6-7}
		& 3-NN & 6-NN & 3-NN & 6-NN & 3-NN & 6-NN \\
		\midrule
		Best Matrices & \textbf{0.922}  & \textbf{0.963}  & \textbf{0.986}  & \textbf{0.998}  & \textbf{0.993}  & \textbf{0.991}  \\
		& $\bar{\mathbf{P}}_{0.6,0.0}(1)$ & $\bar{\mathbf{P}}_{0.1,0.9}(1)$ & $\bar{\mathbf{P}}_{0.2,0.8}(2)$ & $\bar{\mathbf{P}}_{0.0,1.0}(2)$ & $\bar{\mathbf{P}}_{0.6,1.0}(2)$ & $\bar{\mathbf{P}}_{0.0,0.8}(2)$ \\
		DF1 & 0.818 & 0.941 & 0.946 & 0.979 & 0.932 & 0.953 \\
		DF2 & 0.854 & 0.935 & 0.930 & 0.986 & 0.976 & 0.956 \\
		GFT & 0.714 & 0.387 & 0.875 & 0.898 & 0.668 & 0.848 \\
		SP2 & 0.621 & 0.432 & 0.915 & 0.948 & 0.585 & 0.915 \\
		SP3 & 0.672 & 0.432 & 0.922 & 0.563 & 0.644 & 0.869 \\
		MRK & 0.689 & 0.709 & 0.805 & 0.809 & 0.615 & 0.770 \\
		\bottomrule
		\label{table 1}
	\end{tabular}
\end{table*}

\begin{figure}[!t]
	\centering
	\subfloat{\includegraphics[width=1.7in]{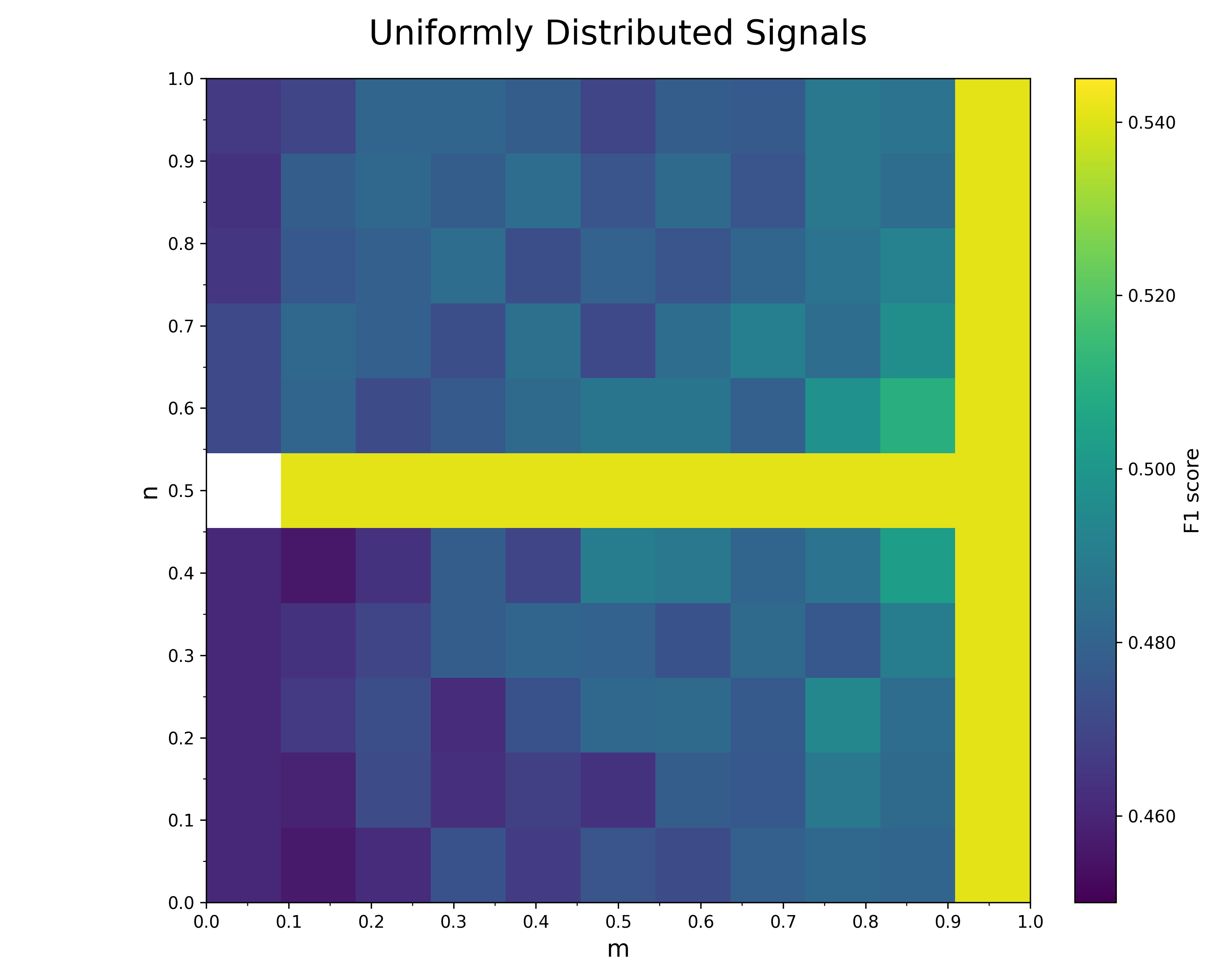}}
	\subfloat{\includegraphics[width=1.7in]{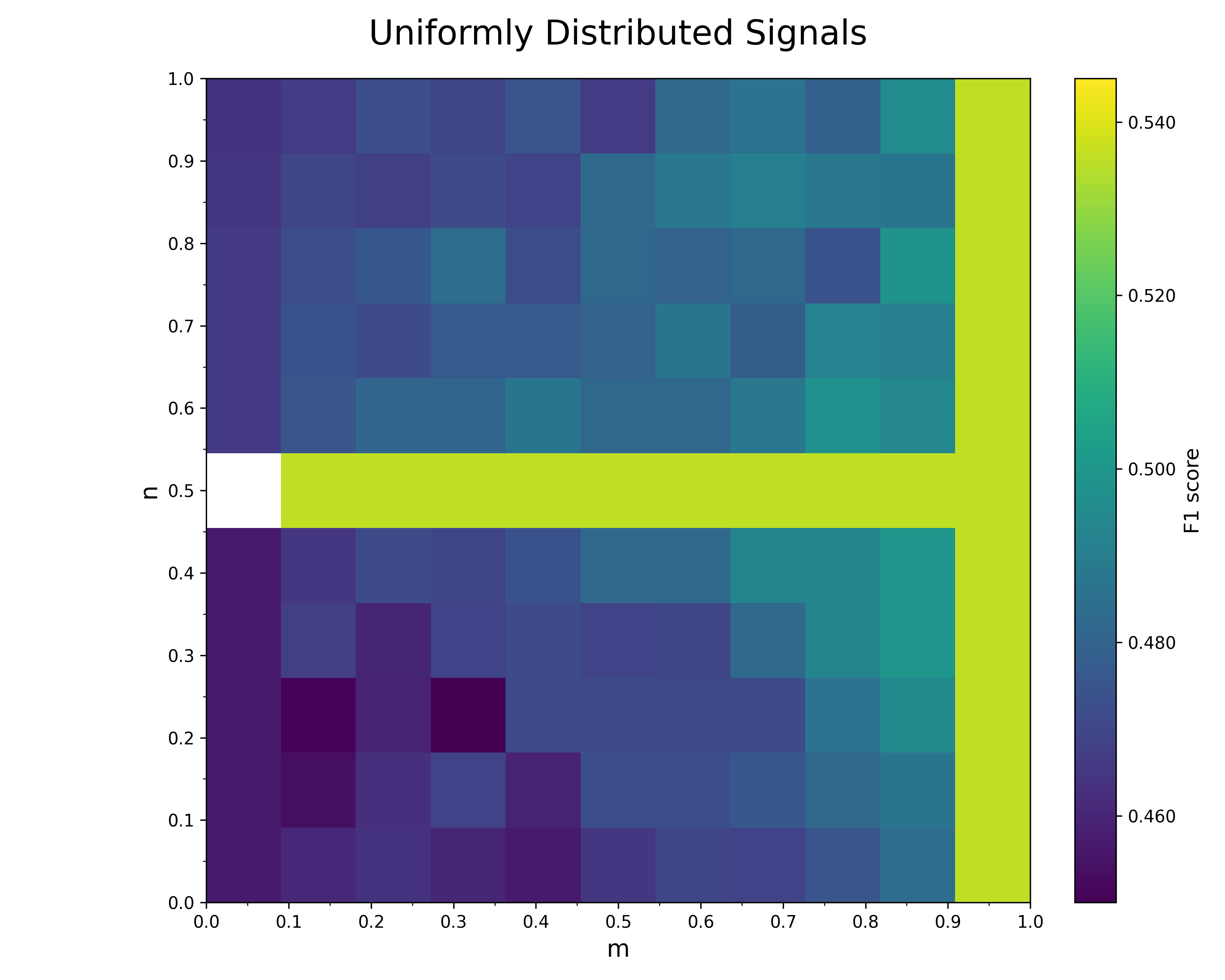}}
	\caption{Figures (Left: $t=1$/Right: $t=2$) depict F1 scores for Uniformly Distributed Signals and for different values of $m$ and $n$ defining $\bar{\mathbf{P}}_{m,n}(t)$ with $N=30$, $k=3$ and $t \in \{1, 2\}$.}
	\label{fig 3}
\end{figure}

\textit{2) Experiment 2:} Uniformly Distributed Signals - sensor malfunction: Consider now a network with $N$ sensors that measure healthy signals $\mathbf{x} \sim \mathcal{U}(-15 \cdot \mathbf{1}, 15 \cdot \mathbf{1})$. The anomaly is described by $b_{\text{max}} = 4$, noise variance equal to 1, and up to 2 anomalous sensors. For each case, both the training and test sets consisted of 200 samples, 100 of which are anomalous, and experiments are conducted on 50 independent data sets. Fig. \ref{fig 3} demonstrates that the "best" matrices for diffusion scales $t=1$ and $t=2$ are achieved at the same matrix. Table \ref{table 2} presents the distribution of F1 scores across all 50 independent experimental runs. The experimental results show that UEM-GFT obtains a superior performance compared to other GSO methods, and the best detections are obtained at the extended-degree matrix.
\begin{table*} 
	\centering
	\caption{F1 scores of different GSOs under different nodes $N$ and  neighboring nodes $k$ for Uniformly Distributed Signals.}
	\begin{tabular}{@{}l@{\hspace{5.46em}}c@{\hspace{5.46em}}c@{\hspace{5.46em}}c@{\hspace{5.46em}}c@{\hspace{5.46em}}c@{\hspace{5.46em}}c@{}}
		\toprule
		& \multicolumn{2}{c}{$N=10$} & \multicolumn{2}{c}{$N=30$} & \multicolumn{2}{c}{$N=50$} \\
		\cmidrule(lr){2-3} \cmidrule(lr){4-5} \cmidrule(lr){6-7}
		& 3-NN & 6-NN & 3-NN & 6-NN & 3-NN & 6-NN \\
		\midrule
		Best Matrices & \textbf{0.519} & \textbf{0.526} & \textbf{0.541} & \textbf{0.540} & \textbf{0.540} & \textbf{0.538} \\
		& $\bar{\mathbf{P}}_{1.0,n}(1)$ & $\bar{\mathbf{P}}_{1.0,n}(2)$ & $\bar{\mathbf{P}}_{1.0,n}(1)$ & $\bar{\mathbf{P}}_{1.0,n}(2)$ & $\bar{\mathbf{P}}_{1.0,n}(1)$ & $\bar{\mathbf{P}}_{1.0,n}(1)$ \\
		DF1 & 0.463 & 0.469 & 0.470 & 0.488 & 0.473 & 0.479 \\
		DF2 & 0.461 & 0.472 & 0.467 & 0.468 & 0.474 & 0.470 \\
		GFT & 0.463 & 0.462 & 0.462 & 0.463 & 0.466 & 0.455 \\
		SP2 & 0.461 & 0.461 & 0.463 & 0.471 & 0.474 & 0.447 \\
		SP3 & 0.462 & 0.461 & 0.457 & 0.458 & 0.457 & 0.463 \\
		MRK & 0.456 & 0.450 & 0.456 & 0.458 & 0.435 & 0.464 \\
		\bottomrule
		\label{table 2}
	\end{tabular}
\end{table*}

\subsection{Simulations Over Real Networks}
\begin{figure}[!t]
	\centering
	\subfloat{\includegraphics[width=1.7in]{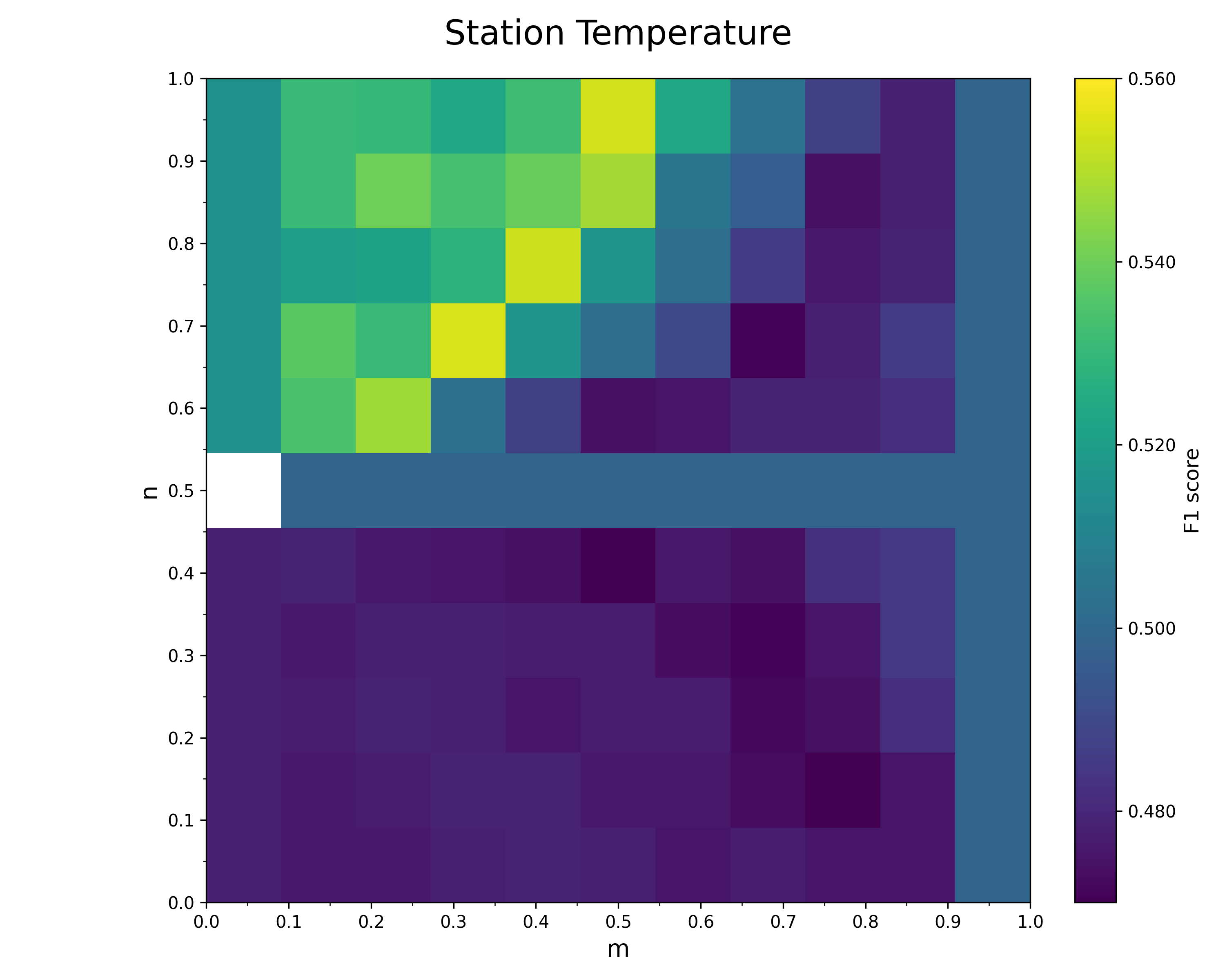}}
	\subfloat{\includegraphics[width=1.7in]{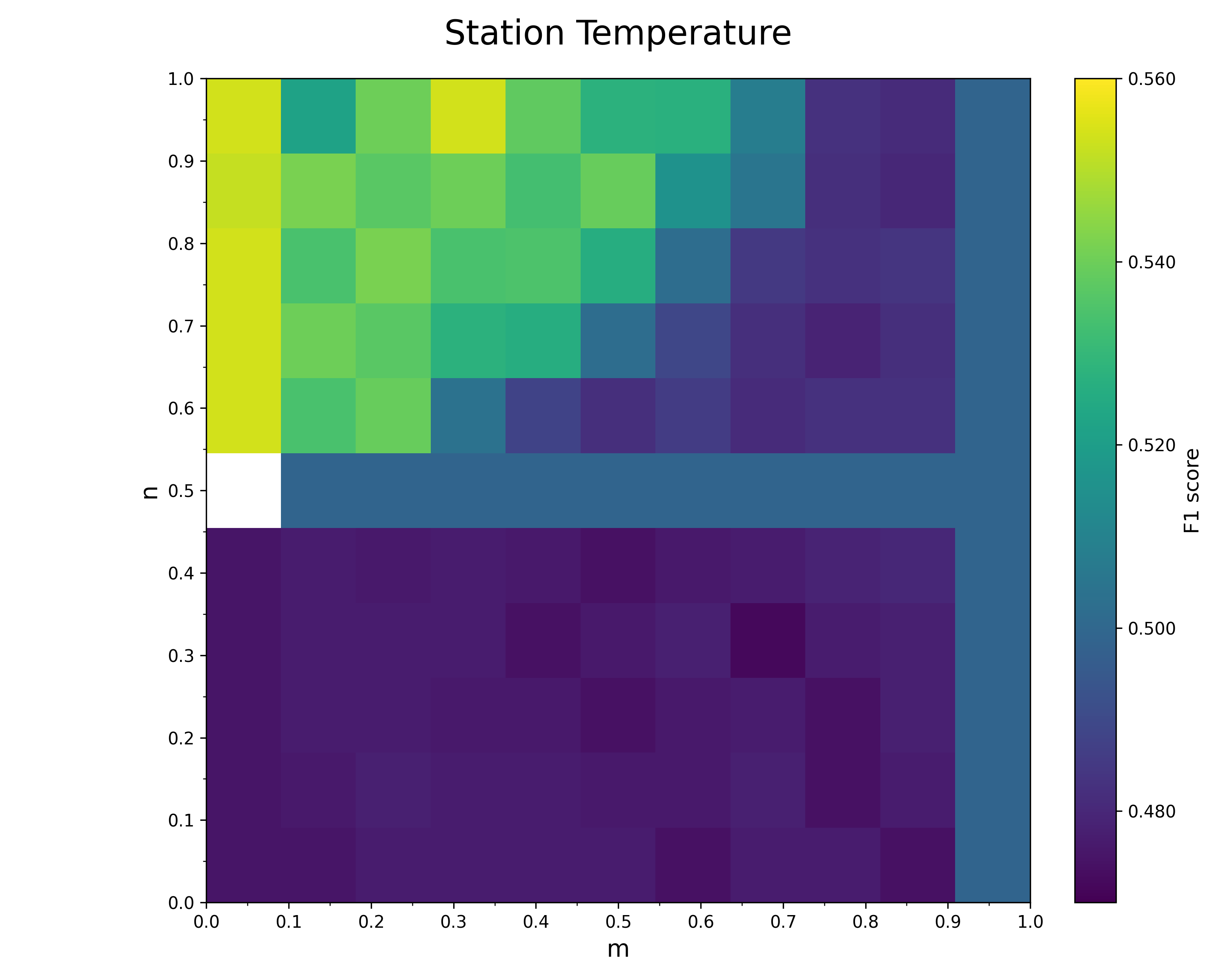}}
	\caption{Figures (Left: $t=1$/Right: $t=2$) depict F1 scores for Station Temperature and for different values of $m$ and $n$ defining $\bar{\mathbf{P}}_{m,n}(t)$ with $N=30$, $k=3$ and $t \in \{1, 2\}$.}
	\label{fig 4}
\end{figure}

\textit{3) Experiment 3:} Station Temperature - sensor malfunction: The database is obtained from the Global Surface Summary of the Day (GSOD) \cite{ref47}, specifically focusing on the geographical region bounded by  $30^\circ\text{N}$ to $49^\circ\text{N}$ latitude and $90^\circ\text{W}$ to $120^\circ\text{W}$ longitude. For our analysis, we process temperature measurements recorded during 2020 from randomly sampled stations within this area, converting all values from Fahrenheit to Celsius. Healthy data range from $-26.1^\circ\text{C}$ to $40.6^\circ\text{C}$. We use $b_{\text{max}} = 5^\circ\text{C}$, noise variance equal to $1 \, ^\circ\text{C}^2$ and up to 5 anomalous sensors.
Sample data are available daily, totaling 366 signals. For each independent experiment, 350 samples are randomly selected, half of which are labeled as anomalies. Subsequently, these 350 samples were divided equally into a training set and a test set. As shown in Fig. \ref{fig 4}, the "maps" illustrating F1 scores vary with different diffusion scales. Results for the Station Temperature are presented in Table \ref{table 3}. The results show that there are cases where sGFT detection is not as effective as using traditional Laplacian and shortest path based GSOs, but with our proposed method it obtains the best detection results.
\begin{table*} 
	\centering
	\caption{F1 scores of different GSOs under different nodes $N$ and  neighboring nodes $k$ for Station Temperature.}
	\begin{tabular}{@{}l@{\hspace{5em}}c@{\hspace{5em}}c@{\hspace{5em}}c@{\hspace{5em}}c@{\hspace{5em}}c@{\hspace{5em}}c@{}}
		\toprule
		& \multicolumn{2}{c}{$N=10$} & \multicolumn{2}{c}{$N=30$} & \multicolumn{2}{c}{$N=50$} \\
		\cmidrule(lr){2-3} \cmidrule(lr){4-5} \cmidrule(lr){6-7}
		& 3-NN & 6-NN & 3-NN & 6-NN & 3-NN & 6-NN \\
		\midrule
		 Best Matrices & \textbf{0.587} & \textbf{0.588} & \textbf{0.555} & \textbf{0.588} & \textbf{0.537} & \textbf{0.561} \\
		& {$\bar{\mathbf{P}}_{0.3,0.8}(2)$} & {$\bar{\mathbf{P}}_{0.2,0.8}(2)$} & {$\bar{\mathbf{P}}_{0.3,0.7}(1)$} & {$\bar{\mathbf{P}}_{0.3,0.8}(2)$} & {$\bar{\mathbf{P}}_{0.1,1.0}(2)$} & {$\bar{\mathbf{P}}_{0.3,1.0}(2)$} \\
		DF1 & 0.563 & 0.567 & 0.554 & 0.567 & 0.516 & 0.535 \\
		DF2 & 0.578 & 0.565 & 0.528 & 0.573 & 0.511 & 0.542 \\
		GFT & 0.567 & 0.575 & 0.519 & 0.561 & 0.511 & 0.506 \\
		SP2 & 0.534 & 0.582 & 0.534 & 0.574 & 0.496 & 0.529 \\
		SP3 & 0.548 & 0.582 & 0.551 & 0.585 & 0.509 & 0.519 \\
		MRK & 0.478 & 0.481 & 0.469 & 0.471 & 0.473 & 0.475 \\
		\bottomrule
		\label{table 3}
	\end{tabular}
\end{table*}

\begin{figure}[!t]
	\centering
	\subfloat{\includegraphics[width=1.7in]{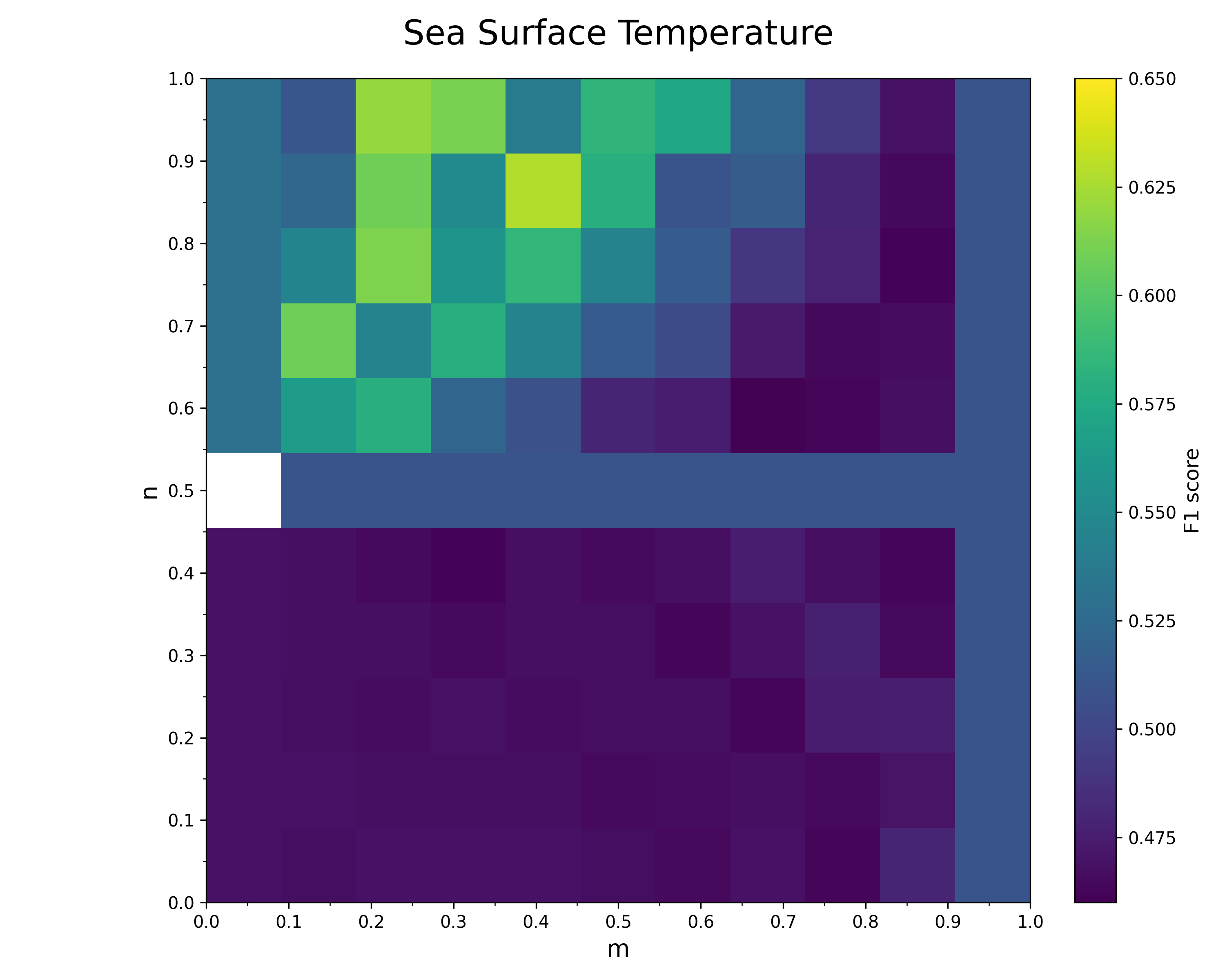}}
	\subfloat{\includegraphics[width=1.7in]{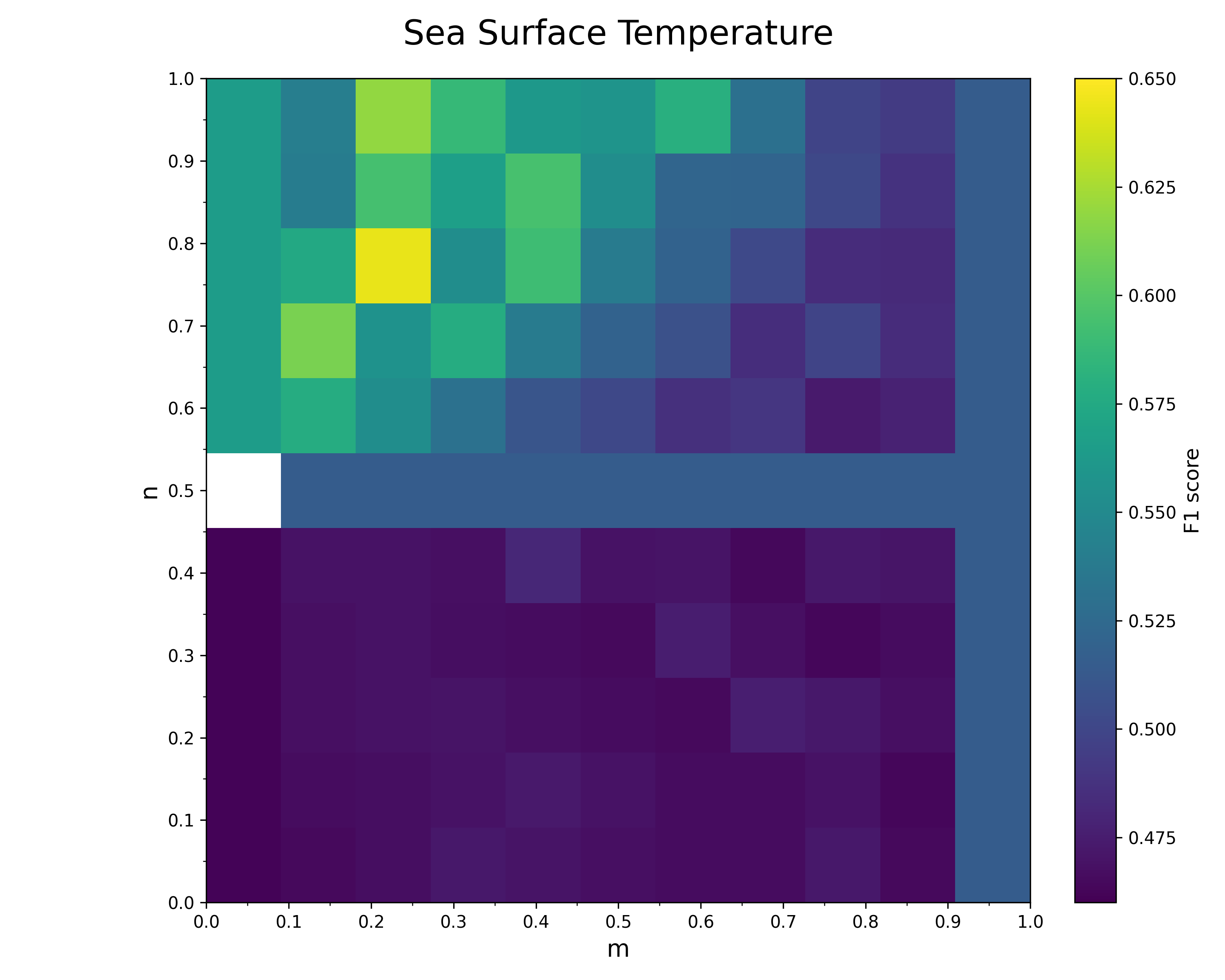}}
	\caption{Figures (Left: $t=1$/Right: $t=2$) depict F1 scores for SST and for different values of $m$ and $n$ defining $\bar{\mathbf{P}}_{m,n}(t)$ with $N=30$, $k=3$ and $t \in \{1, 2\}$.}
	\label{fig 5}
\end{figure}

\textit{4) Experiment 4:} Sea Surface Temperature (SST) \cite{ref48} - sensor malfunction: The SST dataset consists of the monthly captured sea surface temperatures. Healthy data range from $0.02^\circ\text{C}$ to $30.72^\circ\text{C}$. The anomaly is described by $b_{\text{max}} = 4^\circ\text{C}$, noise variance equal to $0.6 \, ^\circ\text{C}^2$, and up to 3 anomalous sensors. Each run incorporates the first 500 months  as samples, with training and test datasets subsequently constructed following the methodology established in Experiment 3. The "maps" displaying F1 scores, as depicted in Fig. \ref{fig 5}, differ across varying diffusion scales. Table \ref{table 4} lists the F1 scores attained from 50 separate experimental runs. The results demonstrate that our proposed method achieves superior detection performance on this dataset, while the Markov-matrix-based approach yields the least effective outcomes. 
\begin{table*}
	\centering
	\caption{F1 scores of different GSOs under different nodes $N$ and  neighboring nodes $k$ for SST.}
	\begin{tabular}{@{}l@{\hspace{5em}}c@{\hspace{5em}}c@{\hspace{5em}}c@{\hspace{5em}}c@{\hspace{5em}}c@{\hspace{5em}}c@{}}
		\toprule
		& \multicolumn{2}{c}{$N=10$} & \multicolumn{2}{c}{$N=30$} & \multicolumn{2}{c}{$N=50$} \\
		\cmidrule(lr){2-3} \cmidrule(lr){4-5} \cmidrule(lr){6-7}
		& 3-NN & 6-NN & 3-NN & 6-NN & 3-NN & 6-NN \\
		\midrule
		Best Matrices & \textbf{0.714} & \textbf{0.632} & \textbf{0.644} & \textbf{0.630} & \textbf{0.673} & \textbf{0.647} \\
		& {$\bar{\mathbf{P}}_{0.1,0.8}(2)$} & {$\bar{\mathbf{P}}_{0.7,1.0}(2)$} & {$\bar{\mathbf{P}}_{0.2,0.8}(2)$} & {$\bar{\mathbf{P}}_{0.2,0.7}(1)$} & {$\bar{\mathbf{P}}_{0.3,0.8}(1)$} & {$\bar{\mathbf{P}}_{0.3,0.7}(2)$} \\
		DF1 & 0.569 & 0.576 & 0.584 & 0.608 & 0.610 & 0.585 \\
		DF2 & 0.575 & 0.574 & 0.558 & 0.611 & 0.604 & 0.618 \\
		GFT & 0.573 & 0.542 & 0.536 & 0.609 & 0.615 & 0.599 \\
		SP2 & 0.539 & 0.482 & 0.578 & 0.563 & 0.599 & 0.594 \\
		SP3 & 0.472 & 0.482 & 0.577 & 0.518 & 0.596 & 0.492 \\
		MRK & 0.458 & 0.461 & 0.467 & 0.468 & 0.465 & 0.463 \\
		\bottomrule
		\label{table 4}
	\end{tabular}
\end{table*}

\begin{figure}[!t]
	\centering
	\subfloat{\includegraphics[width=1.7in]{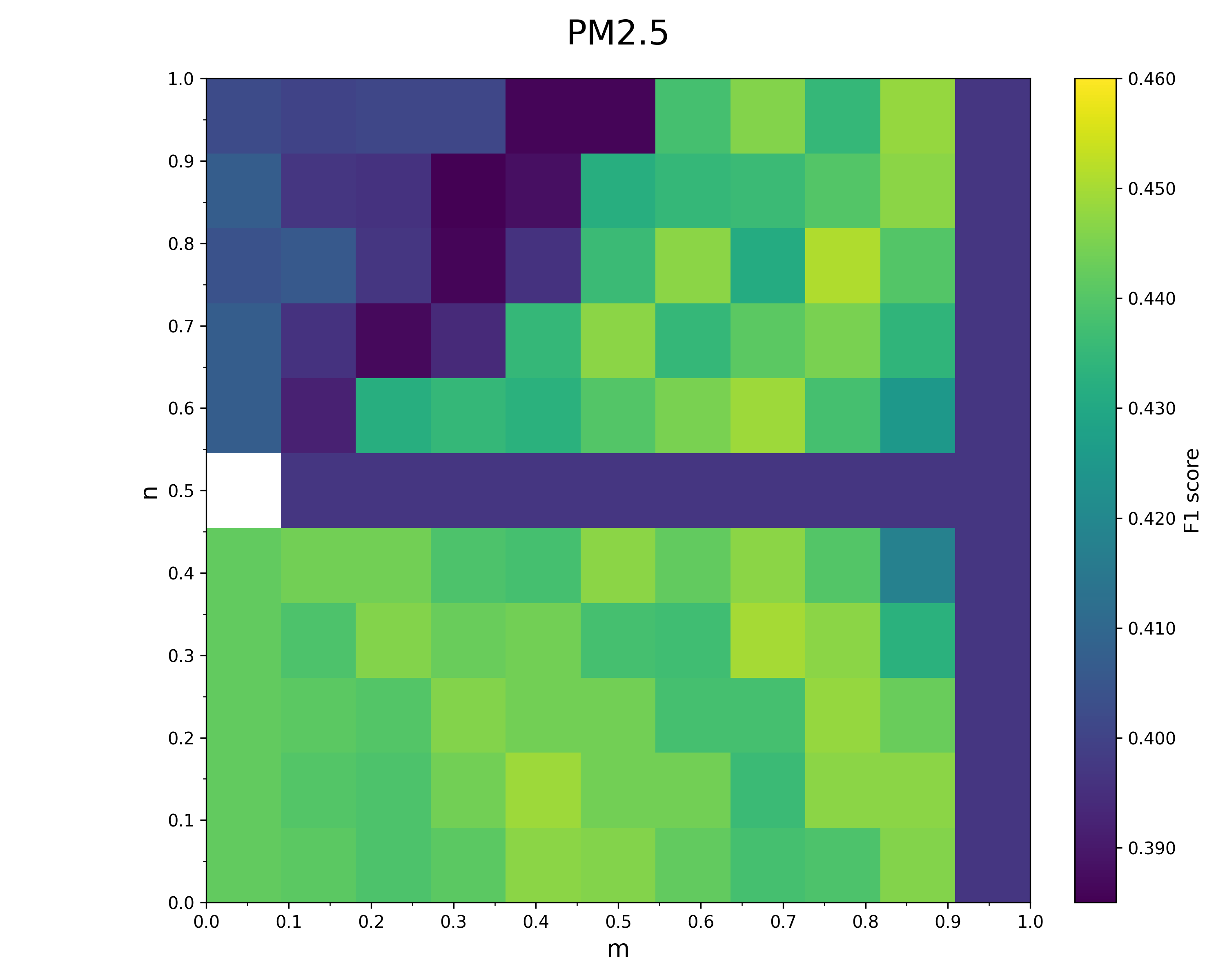}}
	\subfloat{\includegraphics[width=1.7in]{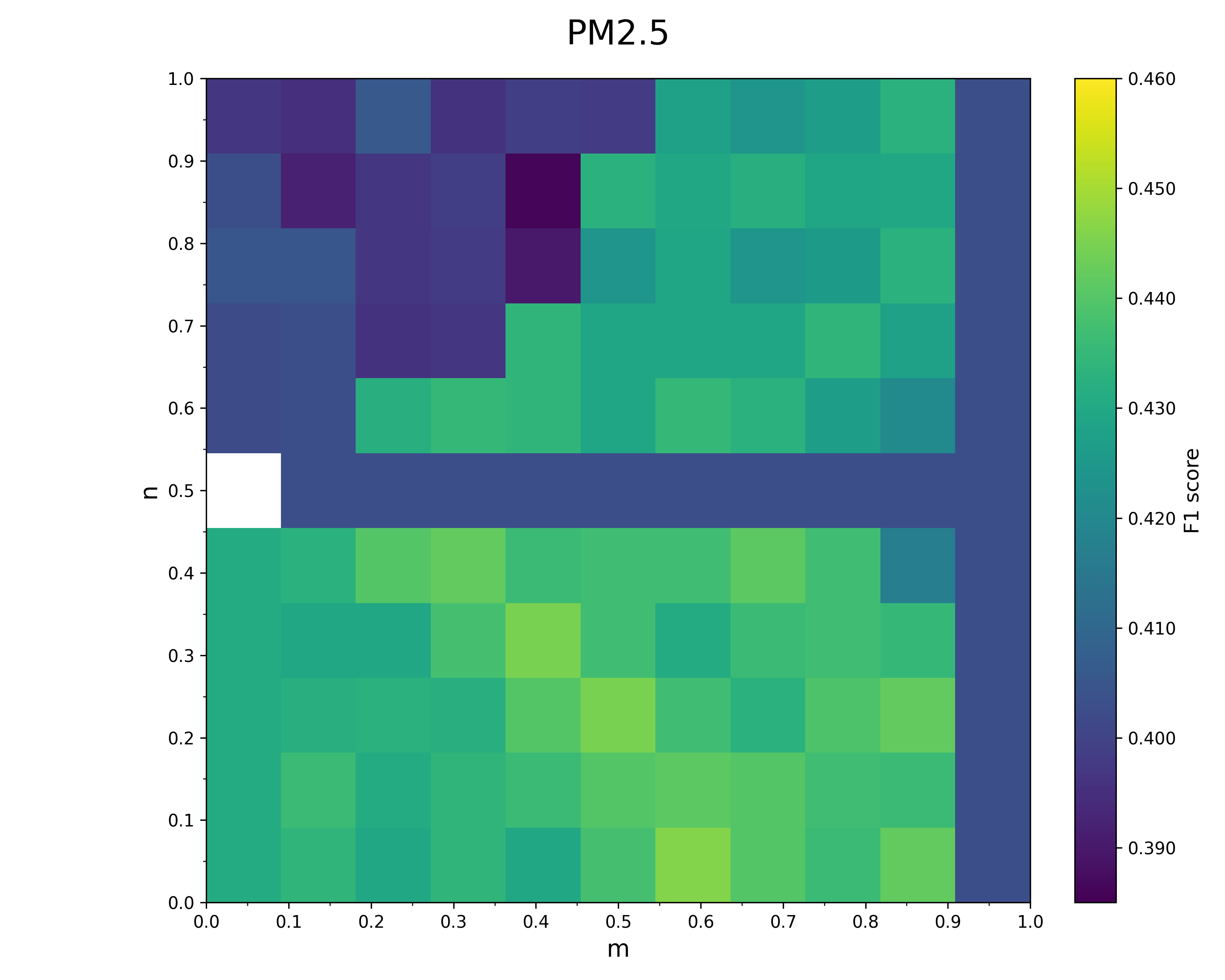}}
	\caption{Figures (Left: $t=1$/Right: $t=2$) depict F1 scores for PM2.5 and for different values of $m$ and $n$ defining $\bar{\mathbf{P}}_{m,n}(t)$ with $N=30$, $k=3$ and $t \in \{1, 2\}$.}
	\label{fig 6}
\end{figure}

\textit{5) Experiment 5:} Particulate Matter 2.5 (PM2.5) \cite{ref48} - sensor malfunction: The PM2.5 consists of the daily mean particle matter 2.5 concentration values, where the data is captured daily from sensors in California during the year 2015. Healthy data range from $0\,\mu\text{g}/\text{m}^3$ to $102.7\,\mu\text{g}/\text{m}^3$. We use $b_{\text{max}} = 3\,\mu\text{g}/\text{m}^3$, noise variance equal to $0.8\ (\mu\text{g}/\text{m}^3)^2$ and up to 2 anomalous sensors. The database provides 238 available samples. Each independent experiment involves randomly acquiring 220 samples, with anomalies introduced to half of them. Training and test datasets are constructed following the methodology of Experiment 4. Fig. \ref{fig 6} shows that "maps" depicting F1 scores show distinct patterns at varying diffusion scales. In Table \ref{table 5}, results show that although the detection performance of the Markov-matrix-based approach is comparable or even better than that of sGFT, the UEM-GFT method demonstrates the best detection results.
\begin{table*} 
	\centering
	\caption{F1 scores of different GSOs under different nodes $N$ and  neighboring nodes $k$ for PM2.5.}
	\begin{tabular}{@{}l@{\hspace{5em}}c@{\hspace{5em}}c@{\hspace{5em}}c@{\hspace{5em}}c@{\hspace{5em}}c@{\hspace{5em}}c@{}}
		\toprule
		& \multicolumn{2}{c}{$N=10$} & \multicolumn{2}{c}{$N=30$} & \multicolumn{2}{c}{$N=50$} \\
		\cmidrule(lr){2-3} \cmidrule(lr){4-5} \cmidrule(lr){6-7}
		& 3-NN & 6-NN & 3-NN & 6-NN & 3-NN & 6-NN \\
		\midrule
		Best Matrices & \textbf{0.481} & \textbf{0.450} & \textbf{0.451} & \textbf{0.457} & \textbf{0.456} & \textbf{0.455} \\
		& {$\bar{\mathbf{P}}_{1.0,n}(2)$} & {$\bar{\mathbf{P}}_{0.7,0.8}(1)$} & {$\bar{\mathbf{P}}_{0.8,0.8}(1)$} & {$\bar{\mathbf{P}}_{0.2,0.4}(2)$} & {$\bar{\mathbf{P}}_{0.7,0.3}(1)$} & {$\bar{\mathbf{P}}_{0.5,0.2}(1)$} \\
		DF1 & 0.452 & 0.447 & 0.386 & 0.426 & 0.440 & 0.438 \\
		DF2 & 0.449 & 0.443 & 0.398 & 0.417 & 0.434 & 0.426 \\
		GFT & 0.449 & 0.420 & 0.398 & 0.419 & 0.415 & 0.426 \\
		SP2 & 0.434 & 0.394 & 0.361 & 0.363 & 0.370 & 0.334 \\
		SP3 & 0.429 & 0.394 & 0.397 & 0.362 & 0.358 & 0.363 \\
		MRK & 0.443 & 0.428 & 0.432 & 0.434 & 0.453 & 0.453 \\
		\bottomrule
		\label{table 5}
	\end{tabular}
\end{table*}
\section{Conclusion}
In this paper, we propose the UEM, a novel parametric graph representation that integrates the strengths of the extended adjacency matrix and unified graph representation matrix. The UEM offers a flexible framework for capturing both local and global dependencies in graph-structured data, advancing capabilities of GSP. We theoretically proved its positive semi-definiteness under specific parameter conditions and eigenvalue monotonicity with parameter variations which is validated through simulation on sensor networks. Leveraging UEM, we introduce UEM-GFT, a generalized GFT with tunable spectral parameters for enhanced adaptability. Extensive experiments on synthetic and real-world datasets show that UEM-GFT outperforms existing methods based on extended-Laplacian matrix, Laplacian matrix and other GSOs, achieving higher F1 scores in anomaly detection across diverse signal types and graph structures.

\end{document}